\documentclass[twoside,english]{article}

\usepackage[accepted]{aistats2020}
\usepackage{times}      
\usepackage{helvet}     
\usepackage{courier}    
\usepackage{url}        
\usepackage{graphicx}   
\usepackage{authblk}

\usepackage{enumitem}
\usepackage{amsmath}
\usepackage{subcaption}

\usepackage{amssymb,amsmath,amsthm}
\usepackage{multicol}
\usepackage{bbm}
\usepackage{microtype}
\usepackage{booktabs} 
\usepackage{algorithm}
\usepackage{algorithmic}
\usepackage{xcolor}
\usepackage{multirow}
\usepackage{inputenc}
\usepackage{babel}
\usepackage[title]{appendix}

\graphicspath{{image/}}

\newcommand{\D}{\mathcal{D}}
\newcommand{\Q}{\mathcal{Q}}
\newcommand{\Proba}{\mathbb{P}}
\newcommand{\R}{\mathbb{R}}

\newcommand{\E}{\mathbb{E}}

\newcommand{\btheta}{\boldsymbol{\theta}}
\newcommand{\calH}{\mathcal{H}}
\newcommand{\calX}{\mathcal{X}}

\newcommand{\calY}{\mathcal{Y}}

\newcommand{\Rad}{\mathrm{Rad}}

\newtheorem{theorem}{Theorem}
\newtheorem{lemma}{Lemma}
\newtheorem{corollary}{Corollary}
\newtheorem{definition}{Definition}

\begin{document}

%

%
\runningauthor{Changjian Shui, Fan Zhou, Christian Gagn\'e, Boyu Wang}

\twocolumn[
\aistatstitle{Deep Active Learning: Unified and Principled Method for Query and Training}
\aistatsauthor{Changjian Shui$^1$ \And Fan Zhou$^1$ \And  Christian Gagn\'e$^{1,2}$ \And Boyu Wang$^{3,4}$}
\aistatsaddress{$^1$Universit\'e Laval\quad $^2$Mila, Canada CIFAR AI Chair\quad $^3$University of Western Ontario\quad $^4$Vector Institute} 
]

\begin{abstract}
In this paper, we are proposing a unified and principled method for both the querying and training processes in deep batch active learning. We are providing theoretical insights from the intuition of modeling the interactive procedure in active learning as distribution matching, by adopting the Wasserstein distance. As a consequence, we derived a new training loss from the theoretical analysis, which is decomposed into optimizing deep neural network parameters and batch query selection through alternative optimization. In addition, the loss for training a deep neural network is naturally formulated as a min-max optimization problem through leveraging the unlabeled data information. Moreover, the proposed principles also indicate an \emph{explicit} uncertainty-diversity trade-off in the query batch selection. Finally, we evaluate our proposed method on different benchmarks, consistently showing better empirical performances and a better time-efficient query strategy compared to the baselines.
\end{abstract}

\section{Introduction}
Deep neural networks (DNNs) achieved unprecedented success for many supervised learning tasks such as image classification and object detection. Although DNNs are successful in many scenarios, there still exists an obvious limitation: the requirement for a large set of labeled data. To address this issue, \emph{Active Learning} (AL) appears as a compelling solution by searching the most informative data points (batch) to label from a pool of unlabeled samples in order to maximize prediction performance.

\begin{figure}[!t]
    \centering 
	\includegraphics[width=0.48\textwidth]{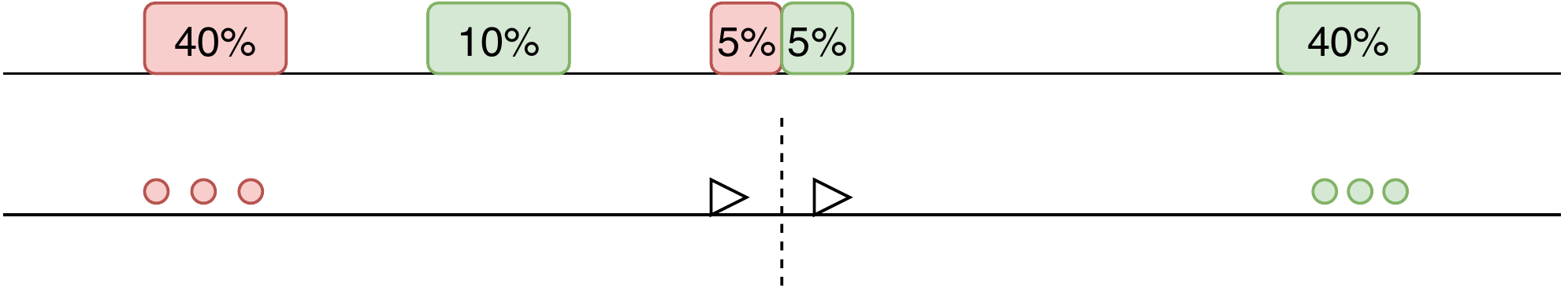}
	\caption{Sampling bias in AL \cite{dasgupta2011two}. In the one dimensional binary classification problem (prediction red/green), the data generation distribution consists of four uniform intervals. Red/Green dots: the initial observations; dotted line: estimated decision boundary from the initial samples; triangles: querying samples according to the uncertainty based strategies w.r.t.\ current decision boundary.}
\label{fig:active_bias}
\end{figure}

How to search the most informative samples in the context of DNN? A common solution is to apply DNN's output confidence score as an uncertainty acquisition function to conduct the query~\cite{settles2012active,gal2017deep,Haussmann2019active}. However, a well-known issue for uncertainty-based sampling in AL is the so-called \emph{sampling bias}~\cite{dasgupta2011two}: the current labeled points are not representative of the underlying distribution. For example, as shown in Fig.~\ref{fig:active_bias}, let us assume that the very few initial samples we obtain lie in the two extreme regions. Then based on these initial observations, the queried samples nearest to the currently estimated decision boundary will lead to a final sub-optimal risk of $10\%$ instead of the true optimal risk of $5\%$. This will be even more severe in high dimensional and complex datasets, which are common when DNNs are employed.

Recent works have considered obtaining a diverse set of samples for training deep learning with a reduced sampling bias. For example, \cite{sener2018active} constructed the core-sets through solving the $K$-center problem. But the search procedure itself is still computationally expensive as it requires constructing a large distance matrix from unlabeled samples. More importantly, it might not be a proper choice particularly for a large-scale unlabeled pool and a \emph{small} query batch, where it is hard to cover the entire data \cite{ash2019deep}.

Instead of focusing exclusively on either uncertainty or diversity instances when determining the query, following a hybrid strategy can be more appropriate. For example, \cite{yin2017deep} heuristically selected a portion of samples according to the uncertainty score for exploitation and the remaining portion used random sampling for exploration. \cite{ash2019deep} collected samples whose gradients span a diverse set of directions for implicitly considering these two. Since such hybrid strategies empirically showed improved performance, a goal of our paper is to derive the query strategy that explicitly considers the uncertainty-diversity trade-off in a principled way. 

Moreover, in the context of deep AL, the available large set of unlabeled samples may be helpful to construct a good feature representation that would potentially allow to improve performance. In order to further promote better results, the question to answer is how we can additionally design a loss for optimizing DNN's weights that would leverage from the unlabeled samples during the training.
 
To address this question, a promising line of work is to integrate the training with a \emph{deep generative model} which naturally focuses on the unlabeled data information \cite{goodfellow2014generative,kingma2013auto}. Only a few works strode in this direction, notably \cite{sinha2019variational} who empirically adopted a $\beta$-VAE to construct the latent variables. Then they adopted the intuition from \cite{DanielGissin2019} of searching the diverse unlabeled batch for samples that do not look like the labeled samples, through an adversarial training based on the $\calH$-divergence \cite{ben2010theory}. In spite of some good performance, this approach still concentrated on empirically designing the training loss, simply adopting the $\calH$-divergence based query strategy.     
In particular, \emph{the formal justifications still remain elusive}, and \emph{the $\calH$-divergence may not be a proper metric for measuring the diversity of the query batch} (see Fig. \ref{fig:comp_dis}), which will be verified in our paper.

In this paper, we are proposing a \emph{unified and principled} approach for \emph{both} a fast querying and a better training procedure in deep AL, relying on the use of labeled and unlabeled examples. We derived the theoretical analysis through modeling the interactive procedure in AL as \emph{distribution matching} by adopting the Wasserstein distance. We also analytically reveal that the Wasserstein distance is better at capturing the diversity, compared with the most common $\calH$-divergence. From the theoretical result, we derived the loss from the distribution matching, which is naturally decomposed into two stages: optimization of DNN parameters and query batch selection, through alternative optimization. 

For the stage of training DNN, the derived loss indicates a min-max optimization problem by leveraging the unlabeled data. More precisely, this involves a maximization of the critic function so as to distinguish the labeled and unlabeled empirical distributions based on the Wasserstein distance, while the feature extractor function aims, on the contrary, to confound the distributions (minimization of empirical distribution divergence). In the query stage, the loss for batch selection \emph{explicitly} indicates the uncertainty-diversity trade-off. For the uncertainty, we want to find the samples with low prediction confidence over two different interpretations: the highest least prediction confidence score and the uniform prediction score (Section~\ref{section:query}). As for the diversity, we want to find the unlabeled batch holding a larger transport cost w.r.t.\ the labeled set under Wasserstein distance (i.e.\ not looking like the current labelled ones), which has been shown as a good metric for measuring diversity. 

Finally, we tested our proposed method on different benchmarks, showing a consistently improved performance, particularly in the initial training, and a much faster query strategy compared with the baselines.
The results reaffirmed the benefits and potential of deriving unified principles for Deep AL. We also hope it will open up a new avenue for rethinking and designing query efficient and principled Deep AL algorithms in the future.

\section{Active Learning as Distribution Matching}
In supervised learning, observations $\hat{\D}$ are i.i.d.\ generated by the underlying distribution $\D$ and a labeling function $h^{\star}$, i.e.\ $\{(x_i,h^{\star}(x_i))\}_{i=1}^N$ with $x_i\sim \D$. While in AL, the querying sample is not an i.i.d.\ procedure w.r.t.\ $\D$ --- otherwise it will be simple random sampling. Thus we assume in AL that the query procedure is an i.i.d.\ empirical process following a distribution $\Q \neq \D$. For example, in the disagreement based approach, $\Q$ can be somehow regarded as a uniform distribution over the disagreement region. Then the interactive procedure can be viewed as estimating a proper $\Q$ to control the generalization error w.r.t.\ $(\D,h^{\star})$. 

\subsection{Preliminaries}
We define the hypothesis $h\in\calH:\calX\to\calY$ over $\calX\subseteq\R^d$ and $\calY\in[0,1]$, and loss function $\ell: \calY\times\calY\to \R^{+}$. The expected risk w.r.t.\ $\D$ is $R_{\D}(h) = \E_{x\sim\D} \ell(h(x),h^{\star}(x))$ and empirical risk $\hat{R}_{\D}(h) = \frac{1}{N}\sum_{i=1}^N \ell(h(x_i),y_i)$. $\Proba(\calX)$ is the set of all probability measures over $\calX$.  We assume that the loss $\ell$ is symmetric, $L$-Lipschitz and $M$-upper bounded and $\forall h\in\calH$ is at most $H$-Lipschitz function. 

\paragraph{Wasserstein Distance} Given two probability measures $\D\in\Proba(\calX)$ and $\Q\in\Proba(\calX)$, the \emph{optimal transport} (or Monge-Kantorovich)
problem can be defined as searching for a probabilistic coupling (joint probability distribution)
$\gamma\in\Proba(\Omega\times\Omega)$ for $x_{\D}\sim\D$ and $x_{\Q}\sim\Q$ that are minimizing the
cost of transport w.r.t.\ some cost function $c$:
\begin{align*}
& \mathrm{argmin}_{\gamma} \int_{\calX\times\calX} c(x_{\D},x_{\Q})^p d\gamma(x_{\D},x_{\Q}),\\
& \mathrm{s.t.}\quad \mathbf{P}^{+} \#\gamma = \D;\quad \mathbf{P}^{-} \#\gamma = \Q,
\end{align*}
where $\mathbf{P}^{+}$ and $\mathbf{P}^{-}$ is the marginal projection over $\Omega\times\Omega$  and $\#$ denotes the push-forward measure. The $p$-Wasserstein distance between $\D$ and $\Q$ for any $p \geq 1$ is defined as:
\begin{equation*}
W_p^p(\D,\Q) = \inf_{\gamma\in \Pi(\D,\Q)} \int_{\calX \times \calX} c(x_{\D},x_{\Q})^p d \gamma(x_{\D},x_{\Q}),
\end{equation*}
where $c:\calX\times\calX \to \R^{+}$ is the cost function of transportation of one unit of mass $x$ to $y$ and $\Pi(\D,\Q)$ is the collection of all joint probability measures on $\calX \times \calX$ with marginals $\D$ and $\Q$. Throughout this paper, we only consider the case of $p=1$, i.e.\ the Wasserstein-1 distance and the cost function as Euclidean ($\ell_2$) distance. 

\paragraph{Labeling Function Assumption} Some theoretical works show that AL cannot improve the sample complexity in the worst case, thus identifying properties of the AL paradigm is beneficial \cite{urner2013probabilistic}. For example, \cite{urner2013plal} defined a formal \emph{Probabilistic Lipschitz} condition, in which the Lipschitzness condition is relaxed and formalizes the intuition that \emph{under suitable feature representation, the probability of two close points having different labels is small} \cite{urner2013probabilistic}. We adopt the Joint Probabilistic Lipschitz property, which can be viewed as an extension of \cite{pentina2018multi} and is also coherent with \cite{NIPS2017_6963}.

\begin{definition}
Let $\phi:\R\to[0,1]$. We say labeling function $h^{\star}$ is $\phi(\lambda)$-$(\D,\Q)$ Joint Probabilistic Lipschitz if $\mathrm{supp}(\Q)\subseteq\mathrm{supp}(\D)$ and for all $\lambda>0$ and all distribution coupling $\gamma\in\Pi(\D,\Q)$:
\begin{equation}
     \Proba_{(x_\D,x_\Q)\sim\gamma}[|h^{\star}(x_\D)-h^{\star}(x_\Q)|>\lambda\|x_\D-x_\Q\|_2]\leq \phi(\lambda),
     \label{proba_lip}
\end{equation}
where $\phi(\lambda)$ reflects the decay property. \cite{urner2013plal} showed that the faster the decay of $\phi(\lambda)$ with $\lambda\to 0$, the better the labeling function and
the easier it is to learn the task.
\end{definition}

\subsection{Bound related Querying Distribution}
In this part, we will derive the relation between the querying and the data generation distribution.
\begin{theorem}
Supposing $\D$ is the data generation distribution and $\Q$ is the querying distribution. If the loss $\ell$ is symmetric, $L$-Lipschitz; $\forall h\in\calH$ is at most $H$-Lipschitz function and the underlying labeling function $h^{\star}$ is $\phi(\lambda)$-$(\D,\Q)$ Joint Probabilistic Lipschitz; then the expected risk w.r.t.\ $\D$ can be upper bounded by:
\begin{equation}
    R_\D(h) \leq R_{\Q}(h) + L(H+\lambda) W_1(\D,\Q)  +  L\phi(\lambda).
    \label{transfer_risk}
\end{equation}
\end{theorem}
See the proof in the supplementary material. From Eq.~(\ref{transfer_risk}), the expected risk of $\D$ is upper bounded by the expected risk w.r.t. the query distribution $\Q$, the Wasserstein distance $W_1(\D,\Q)$, and the labeling function property $\phi(\lambda)$. That means a desirable query should hold a small expected risk with a better matching the original distribution $\D$ (diversity). 

\paragraph{Non-Asymptotic Analysis} Moreover, we can extend the non-asymptotic analysis of Theorem 1 since we generally have finite observations. The proof is also provided in the supplementary material.
\begin{corollary}
Supposing we have the finite observations which are i.i.d.\ generated from $\D$ and $\Q$: $\hat{D}=\frac{1}{N}\sum_{i=1}^N \delta\{x_{\D}^i\}$ and $\hat{Q}=\frac{1}{N_q}\sum_{i=1}^{N_q} \delta\{x_{\Q}^i\}$ with $N_{q}\leq N$. Then with probability $\geq 1-\delta$, the expected risk w.r.t.\ $\D$ can be further upper bounded by:
\begin{equation*}
\begin{split}
     R_\D(h) & \leq \hat{R}_{\Q}(h) +  L(H+\lambda) W_1(\hat{\D},\hat{\Q}) \\
             & +  L\phi(\lambda) + 2L \Rad_{N_q}(h) + \kappa(\delta,N,N_q),
\end{split}
\end{equation*}
where $\kappa(\delta,N,N_q) = \mathcal{O}(N^{-1/s_d} + N_q^{-1/s_q} + \sqrt{\frac{\log(1/\delta)}{N}} + \sqrt{\frac{\log(1/\delta)}{N_q}})$ with some positive constants $s_d$ and $s_q$. $\Rad_{N_q}(h) = \E_{S\sim\Q^{N_q}}\E_{\sigma_{1}^{N_q}}[\sup_{h} \frac{1}{N_q}\sum_{i=1}^{N_q} \sigma_i h(x_i)]$ is the expected Rademacher complexity generally with $\Rad_{N_q}(h) = \mathcal{O}(\sqrt{\frac{1}{N_q}})$ (e.g \cite{mohri2018foundations}).
\end{corollary}

\subsection{Why Wasserstein Distance}
In the context of deep active learning, current work such as  \cite{DanielGissin2019,sinha2019variational} generally explicitly or implicitly adopted the idea of $\calH$-divergence \cite{ben2010theory}: $d_{\calH}(\D,\Q)= 1-2\epsilon$, with $\epsilon$ the prediction error when training a binary classifier to \emph{discriminate} the observations sampling from the query and original distribution. Thus a smaller error facilitates the separation of the two distributions with larger $\calH$-divergence and vice versa.

However, we should notice that in AL, $\mathrm{supp}(\Q)\subseteq\mathrm{supp}(\D)$, thus $\calH$-divergence may not be a good metric for indicating the diversity property of the querying distribution. On the contrary, Wasserstein distance reflects an optimal transport cost for moving one distribution to another. A smaller transport cost means a better coverage of the distribution $\D$.

\begin{figure}[!t]
    \centering 
	\includegraphics[width=0.49\textwidth]{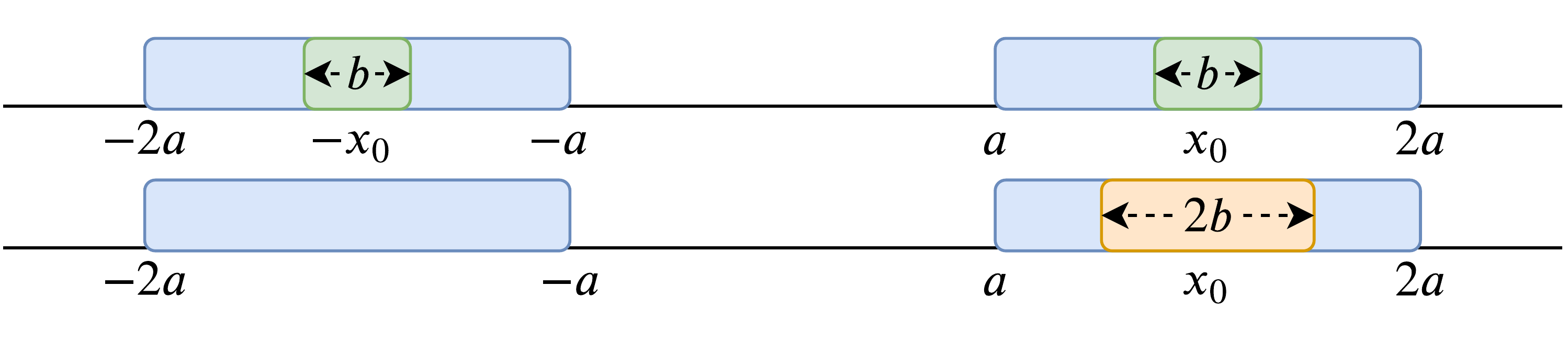}
	\caption{$\calH$-divergence vs.\ Wasserstein distance for $\D$-$\Q$ distribution matching. The desirable query distribution should be more diverse (first row) for avoiding \emph{sampling bias} (second row). The computational result shows that $\calH$-divergence is not a proper metric to measure query diversity while Wasserstein is.}
\label{fig:comp_dis}
\end{figure}

For a better understanding of this problem, we give an illustrative example by computing the exact $\calH$-divergence and Wasserstein-1 distance in one-dimension, shown in Fig.~\ref{fig:comp_dis}. More specifically, we have three uniform distributions: $\D_1$ the original data distribution, $\D_2$, $\D_3$ two different query distributions: 
\begin{align*}
   & \D_1 \sim \mathcal{U}\big([-2a,-a]\cup[a,2a]\big),\\
   & \D_2 \sim \mathcal{U}\big([-x_0-\frac{b}{2},-x_0+\frac{b}{2}] \cup [x_0-\frac{b}{2},x_0+\frac{b}{2}]\big),\\
   & \D_3 \sim \mathcal{U}\big( [x_0-b,x_0+b] \big).
\end{align*}

In AL, we can further assume $\mathrm{supp}(\D_2)\subseteq\mathrm{supp}(\D_1)$, $\mathrm{supp}(\D_3)\subseteq\mathrm{supp}(\D_1)$ and $a>b>0$. For $\calH$-divergence, we set the classifier as a threshold function $f(x)= \mathbf{1}\{x\geq p\}$. Then we can compute the exact $d_{\calH}(\cdot,\cdot)$ and $W_1(\cdot,\cdot)$:
\begin{equation}
    \begin{split}
         d_{\calH}(\D_1,\D_2) & = d_{\calH}(\D_1,\D_3)\\
         \min_{x_0} W_1(\D_1,\D_3) & > \max_{x_0}W_1(\D_1,\D_2)
    \end{split}
    \label{eq:comp_result}
\end{equation}
From Eq.~(\ref{eq:comp_result}), the $\calH$-divergence indicates the same divergence result where Wasserstein-1 distance exactly captures the property of diversity: \emph{more diverse query distribution $\Q$ means smaller Wasserstein-1 distance $W_1(\D,\Q)$}. 

\section{Practical Deep Batch Active Learning}
We have discussed the interactive procedure as the distribution matching and also showed that Wasserstein distance is a \emph{proper} metric for measuring the diversity during distribution matching. Based on the aforementioned analysis, in the batch active learning problem, we have labelled data $\hat{L} = \frac{1}{L}\sum_{i=1}^L \delta\{ x_i^{l}\}$ and its labels $\{y_i^{l}\}_{i=1}^L$, unlabelled data $\hat{U} = \frac{1}{U}\sum_{i=1}^U \delta\{ x_i^{u}\}$ and total distribution $\hat{\D} = \hat{L}\cup\hat{U}$ with partial labels $\{y_i^{l}\}_{i=1}^L$. The goal of AL at each interaction is: 1) find a batch $\hat{B} = \frac{1}{B}\sum_{i=1}^B \delta\{x_i^b\}$ with $x_i^b \in \hat{U}$ during the query; 2) find a hypothesis $h\in\calH$ such that:
\begin{equation}
    \min_{\hat{B},h} \hat{R}_{\hat{L} \cup \hat{B}}(h) + \mu W_1(\hat{\D},\hat{L} \cup \hat{B}).
    \label{AL_loss}
\end{equation}
Eq.(\ref{AL_loss}) follows the principles (upper bound) from Theorem 1 and Corollary 1. Moreover, if we fix the hypothesis $h$, the sampled batch holds the following two requirements simultaneously:
\begin{enumerate}[leftmargin=*]
    \item Minimize the empirical error. We will show later it is related to uncertainty based sampling.
    \item Minimize the Wasserstein-1 distance w.r.t.\ original distribution, which encourages a better distribution matching of $\hat{\D}$. 
\end{enumerate}

\subsection{Min-Max Problem in DNN}
Based on Eq.(\ref{AL_loss}), we can extend the loss to the deep representation learning scenario, since directly estimating the Wasserstein-1 distance through solving optimal transport for complex and large-scale data is still a challenging and open problem. Then inspired by \cite{arjovsky2017wasserstein}, we then adopt the min-max optimizing through training the DNN. Namely, according to Kantorovich-Rubinstein duality, Eq.~(\ref{AL_loss}) can be reformulated as:
\begin{equation}
    \min_{\btheta^f,\btheta^h,\hat{B}} \max_{\btheta^{d}}~ \hat{R}(\btheta^f,\btheta^{h}) + \mu  \hat{E}(\btheta^{f},\btheta^d),
    \label{nn_loss}
\end{equation}
where $\btheta^{f}$, $\btheta^{h}$, $\btheta^{d}$ are parameters corresponding to the \emph{feature extractor}, \emph{task predictor} and \emph{distribution critic}; $\hat{R}$ is the predictor loss and $\hat{E}$ is the adversarial (min-max) loss.

We further denote the \emph{parametric task prediction} function $h(x,y,(\btheta^f,\btheta^h))\equiv h(x,y):\calX\times\calY\to(0,1]$ with $\sum_{y}h(x,y)=1$ and 
the \emph{parametric critic} function $g(x,(\btheta^f,\btheta^d))\equiv g(x):\calX\to[0,1]$ with restricting $g(x)$ to the $1$-Lipschitz function (Kantorovich-Rubinstein theorem). Then each term in Eq.~(\ref{nn_loss}) can be expressed as:
\begin{align*}
    & \hat{R}(\btheta^f,\btheta^{h}) = \E_{(x,y)\sim \hat{L}\cup\hat{B}} \ell(h(x,y)),\\
    & \hat{E}(\btheta^f,\btheta^{d}) = \E_{x\sim \hat{\D}} [g(x)]- \E_{x\sim \hat{L}\cup\hat{B}} [g(x)].
\end{align*}

\subsection{Two-stage Optimization}
Then through some computation, we can decompose Eq.~(\ref{nn_loss}) into three terms:
\begin{equation}
    \begin{split}
       & \min_{\btheta^f,\btheta^h,\hat{B}}\max_{\btheta^{d}}~  \underbrace{\frac{1}{L+B}\sum_{(x,y)\in\hat{L}} \ell(h(x,y))}_\text{Training: Prediction Loss}\\
       & + \underbrace{\mu \big(\frac{1}{L+U}\sum_{x\in\hat{U}} g(x) - (\frac{1}{L+B}-\frac{1}{L+U})\sum_{x\in\hat{L}} g(x) \big)  }_\text{Training: Min-max Loss} \\
       & + \underbrace{\frac{1}{L+B}\sum_{(x,y^{?})\in\hat{B}} \ell(h(x,y^{?})) -  \frac{\mu}{L+B}\sum_{x\in\hat{B}} g(x)}_\text{Query}, 
    \end{split}
    \label{dl_loss}
\end{equation}
where the critic function $g(x)$ is $1$-Lipschitz and $L$, $U$, $B$ are the size of labeled, unlabeled, and query data. $y^{?}$ is called the \emph{agnostic-label}, since it is not available during the query stage. Then from Eq.~(\ref{dl_loss}), each interaction of AL can be naturally decomposed into two stages (optimizing DNN and batch selection), through alternative optimization.

\subsection{Training DNN} 
In the training stage, we used all of the observed data to optimize the neural network parameters:
\begin{equation}
    \begin{split}
        & \min_{\btheta^f,\btheta^h}\max_{\btheta^{d}}~  \frac{1}{L+B}\sum_{(x,y)\in\hat{L}} \ell(h(x,y))\\
        & + \mu \big(\frac{1}{L+U}\sum_{x\in\hat{U}} g(x) - (\frac{1}{L+B}-\frac{1}{L+U})\sum_{x\in\hat{L}} g(x) \big), 
    \end{split}
    \label{nn_train_loss}
\end{equation}
with restricting $g(x)$ to the $1$-Lipschitz function. Instead of only minimizing the prediction error, the proposed approach naturally leveraged the unlabeled data information through a min-max training. 
More intuitively, the critic function $g$ aims to evaluate how probable it is that the sample comes from the labeled or unlabeled parts\footnote{We should point out that this is the high-level intuition. More specifically, the critic parameter $\btheta^d$ of $g$ tries to maximize and the feature parameter $\btheta^f$ of $g$ tries to minimize the adversarial loss according to the Wasserstein metric. Moreover the proposed min-max loss differs from the standard Wasserstein min-max loss since they hold different weights (``bias coefficient'')}. According to the loss, given a fixed $g$, when $g(x)\to 1$ meaning that it is highly probable that the samples come from the unlabeled set $x\in\hat{U}$ and vice versa.  Since $B<U$, thus $\frac{1}{L+B}-\frac{1}{L+U} >0$, means the proposed adversarial loss being always valid.

Based on Eq.~(\ref{nn_train_loss}), we call this framework the Wasserstein Adversarial Active Learning (WAAL) in our deep batch AL. The labeled $\hat{L}$ and unlabeled data $\hat{U}$ pass a common feature extractor, then $\hat{L}$ will be used in the prediction and $\hat{L}$, $\hat{U}$ together will be used in the min-max (adversarial) training. In the practical deep learning, we apply the cross entropy loss: $\ell(x,y) = -\log(h(x,y))$\footnote{Although the cross entropy loss does not satisfy the exact assumptions in the theoretical analysis, our later experiments suggest that the proposed algorithm is very effective for cross-entropy loss.}.

\paragraph{Redundancy Trick}
One can directly apply gradient descent to optimize Eq.~(\ref{nn_train_loss}) on the whole dataset. Actually, we generally apply the mini-batch based SGD approach in training the DNN\footnote{We have referred to this as the \emph{training/mini batch} to avoid any confusion with the querying batch mentioned before.}. While a practical concern during the adversarial training procedure is the unbalanced label and unlabeled data during the training procedure. Thus we propose the \emph{redundancy trick} to solve this concern. For abuse of notation, we denote the unbalanced ratio $\gamma = \frac{U}{L}$ and the query ratio $\alpha = \frac{B}{L}$, with the adversarial loss simplified as:
\begin{equation*}
     \mu^{\prime} \big(\frac{1}{U}\sum_{x\in\hat{U}}\, g(x)- \frac{1}{\gamma}\frac{\gamma-\alpha}{1+\alpha} \frac{1}{L}\sum_{x\in\hat{L}}g(x) \big), 
\end{equation*}
with $\mu^{\prime} = \frac{\gamma}{1+\gamma}\mu$. 

Then following the \emph{redundancy trick} for optimizing the adversarial loss, we keep the same mini-batch size $S$ for labelled and unlabeled observations. Due to the existence of the unbalanced data, we simply conduct a sampling with replacement to construct the training batch for the labeled data, then divided by the unbalanced ratio $\gamma$. For each training batch, the adversarial loss can be rewritten as:
\begin{equation}
    \min_{\btheta^f}\max_{\btheta^d}~\mu^{\prime} \big(\frac{1}{S}\sum_{x\in\hat{U}_S} g(x) - C_0 \frac{1}{S}\sum_{x\in\hat{L}_S} g(x)    \big),
    \label{bias_adv}
\end{equation}
where $\hat{U}_s$ $\hat{L}_s$ are unlabeled and labeled training batch and $C_0 = \frac{1}{\gamma^2}\frac{\gamma-\alpha}{1+\alpha}$ is the ``bias coefficient'' in deep active adversarial training. For example, if there exist $1$K labeled samples, $9$K unlabeled samples and a current query batch budget of $1$K, then we can compute $C_0 \approx 0.05$ so as to control excessive reusing of the labelled dataset.  

\subsection{Query Strategy}\label{section:query}
The second stage over the unlabeled data aims to find a querying batch such that:
\begin{equation}
  \mathrm{argmin}_{\hat{B}\subset\hat{U}} \frac{1}{L+B}\sum_{(x,y^{?})\in\hat{B}} \ell(h(x,y^{?})) -  \frac{\mu}{L+B}\sum_{x\in\hat{B}} g(x),
  \label{query_eq}
\end{equation}
where $y^{?}$ is the agnostic-label.

\paragraph{Agnostic-label upper bound loss indicates uncertainty} 
Since we do not known $y^{?}$ during the query, we can instead optimize an \emph{upper bound} of Eq.~(\ref{query_eq}). In the classification problem with cross entropy loss, suppose that we have $\{1,\dots,K\}$ possible outputs with $\sum_{y\in\{1,\dots,K\}}h(x,y)=1$, then we have upper bounds Eq.~(\ref{sig_worst_ub}) and (\ref{l1_ub}), which both reflect the uncertainty measures with different interpretations. 
\begin{enumerate}[leftmargin=*]
    \item Minimizing over the single worst case upper bound indicates the sample with the \emph{highest least prediction confidence score}:
    \begin{equation}
        \min_{x} \ell(h(x,y^{?})) \leq \min_{x}\max_{y\in\{1,\dots,K\}}-\log(h(x,y)).
        \label{sig_worst_ub}
    \end{equation}
    For example, we have two samples with a binary decision score $h(x_1,\cdot)=[0.4,\,0.6]$ and $h(x_2,\cdot)=[0.3,\,0.7]$. Since $\max_{y} -\log(h(x_1,y)) < \max_{y} -\log(h(x_2,y))$, we will choose $x_1$ as the query since the least prediction label confidence $0.4$ is higher. Intuitively such a sample seems uncertain since the least label prediction confidence is high\footnote{In the binary classification problem, it recovers the least prediction
    confidence score approach (Baseline 2), which is a common strategy in AL.}.
    
    \item Minimizing over $\ell_1$ norm upper bound indicates the sample with a \emph{uniformly of prediction confidence score}:
     \begin{equation}
        \min_{x}\ell(h(x,y^{?})) \leq \min_{x}\sum_{y\in\{1,\dots,K\}}-\log(h(x,y)).
        \label{l1_ub}
    \end{equation}
    Intuitively if the sample's prediction confidence trend is more uniform, the more uncertain the sample will be. We can also show the $\min$ arrives when the output score is uniform, as shown in the supplementary material. 
\end{enumerate}

We would like to point out that the upper bounds proposed in Eq.~(\ref{sig_worst_ub},\ref{l1_ub}) are additive, i.e.\ we can apply any convex combination for these two losses as the hybrid uncertain query strategy.

\paragraph{Critic output indicates diversity} As for the critic function $g(x):\calX\to[0,\,1]$ from the adversarial loss, if the critic function output trends to $g(x)\to 1$, it means $x\in\hat{U}$ and vice versa. Then according to the query loss,  we want to select the batch with higher critic values $g(x)$, meaning they look more different than the labelled samples under the Wasserstein metric. 

If the unlabeled samples look like the labeled ones (small $g(x)$ with $x\in\hat{U}$), then under some proper conditions (such as \emph{Probabilistic Lipschitz Condition} in def.~\ref{proba_lip}), such examples can be more easily predicted because we can infer them from their very near neighbours' information. 

On the contrary, the unlabeled samples with high $g(x)$ under the current assumption cannot be effectively predicted by the current labeled data (far away data). Moreover, the $g(x)$ is trained through Wasserstein distance based loss, shown as a proper metric for measuring the diversity. Therefore the query batch with higher critic value ($g(x)$) means a larger transport cost from the labeled samples, indicating that it is more informative and represents diversity. 

\paragraph{Remark} The aforementioned two terms in the query strategy indicate an explicit \emph{uncertainty} and \emph{diversity} trade-off. Uncertainty criteria can reduce the empirical risk but leading to a potential sampling bias. While the diversity criteria can improve the exploration of the distribution while might be inefficient for a small query batch. Our query approach naturally combines these two, for choosing the samples with prediction uncertainty and diversity. 
Moreover, since Eq.~(\ref{query_eq}) is additive, we can easily estimate the query batch through the greedy algorithm. 

\subsection{Proposed Algorithm}
Based on the previous analysis, our proposed algorithm includes a training stage [Eq.~(\ref{nn_train_loss},\ref{bias_adv})] and a query stage [Eq.~(\ref{query_eq}), (\ref{sig_worst_ub}) and (\ref{l1_ub})] for solving Eq.~(\ref{nn_loss}) or (\ref{dl_loss}). We only show the learning algorithm for one interaction in Algorithm \ref{AALN_algo}, then the remaining interactions will be repeated accordingly.

Since the discriminator function $g$ should be restricted in $1$-Lipschitz, we add the gradient penalty term such as \cite{gulrajani2017improved} to $g$ to restrict the Lipschitz property.

\begin{center}
\begin{algorithm}[!t]
		\caption{WAAL: one interaction}
		\begin{algorithmic}[1] 
		\REQUIRE Labeled samples $\hat{L}$, unlabeled samples $\hat{U}$, query budget $B$ and  hyper-parameters (learning rate $\eta$, trade-off rate $\mu$, $\mu^{\prime}$)
        \ENSURE Neural network parameters $\btheta^{f}$, $\btheta^h$, $\btheta^d$ 
        \STATE \(\triangleright\triangleright\triangleright\) \textbf{DNN Parameter Training Stage} \(\triangleleft\triangleleft\triangleleft\)
        \FOR{mini-batch of samples $\{(x^u)\}_{i=1}^S$ from $\hat{U}$}
        \STATE Constructing mini-batch $\{(x^l,y^l)\}_{i=1}^S$ from $\hat{L}$ through sampling with replacement (redundancy trick).
        \STATE Updating $\btheta^h$:  $\btheta^h = \btheta^h - \frac{\eta}{S}\sum_{(x^l,y^l)} \frac{\partial \ell(h((x^l,y^l))}{\partial\btheta^h} $  
        \STATE Updating $\btheta^f$: $\btheta^f = \btheta^f - \frac{\eta}{S} \big(\sum_{(x^l,y^l)} \frac{\partial \ell(h((x^l,y^l))}{\partial\btheta^f} + \mu^{\prime}\{\sum_{x^u} \frac{\partial g(x)}{\partial\btheta^f} - C_0 \sum_{x^l} \frac{\partial g(x)}{\partial\btheta^f}\} \big)$ 
        \STATE Updating $\btheta^d$: $\btheta^d = \btheta^d + \frac{\eta\mu^{\prime}}{S} \{\sum_{x^u} \frac{\partial g(x)}{\partial\btheta^d} - C_0 \sum_{x^l} \frac{\partial g(x)}{\partial\btheta^d}\} $ 
        \ENDFOR
        \STATE \(\triangleright\triangleright\triangleright\) \textbf{Querying Stage} \(\triangleleft\triangleleft\triangleleft\)
		\STATE Applying the convex combination of Eq.~(\ref{sig_worst_ub}) and (\ref{l1_ub}) to compute uncertainly score $\mathcal{U}(x^u)$; \\
		Computing diversity score $g(x^u)$; \\
		Ranking the score $\mathcal{U}(x^u)-\mu g(x^u)$ with $x^u \in\hat{U}$, choosing the smallest $B$ samples, forming querying batch $\hat{B}$  
		\STATE \(\triangleright\triangleright\triangleright\) \textbf{Updating} \(\triangleleft\triangleleft\triangleleft\)
		\STATE $\hat{L} = \hat{L}\cup\hat{B}$, $\hat{U} = \hat{U}\setminus\hat{B}$
        \end{algorithmic}
        \label{AALN_algo}
\end{algorithm}
\end{center}

\begin{figure*}[!t]
\centering
   \begin{subfigure}{0.33\linewidth}
   \centering
   \includegraphics[width=\linewidth]{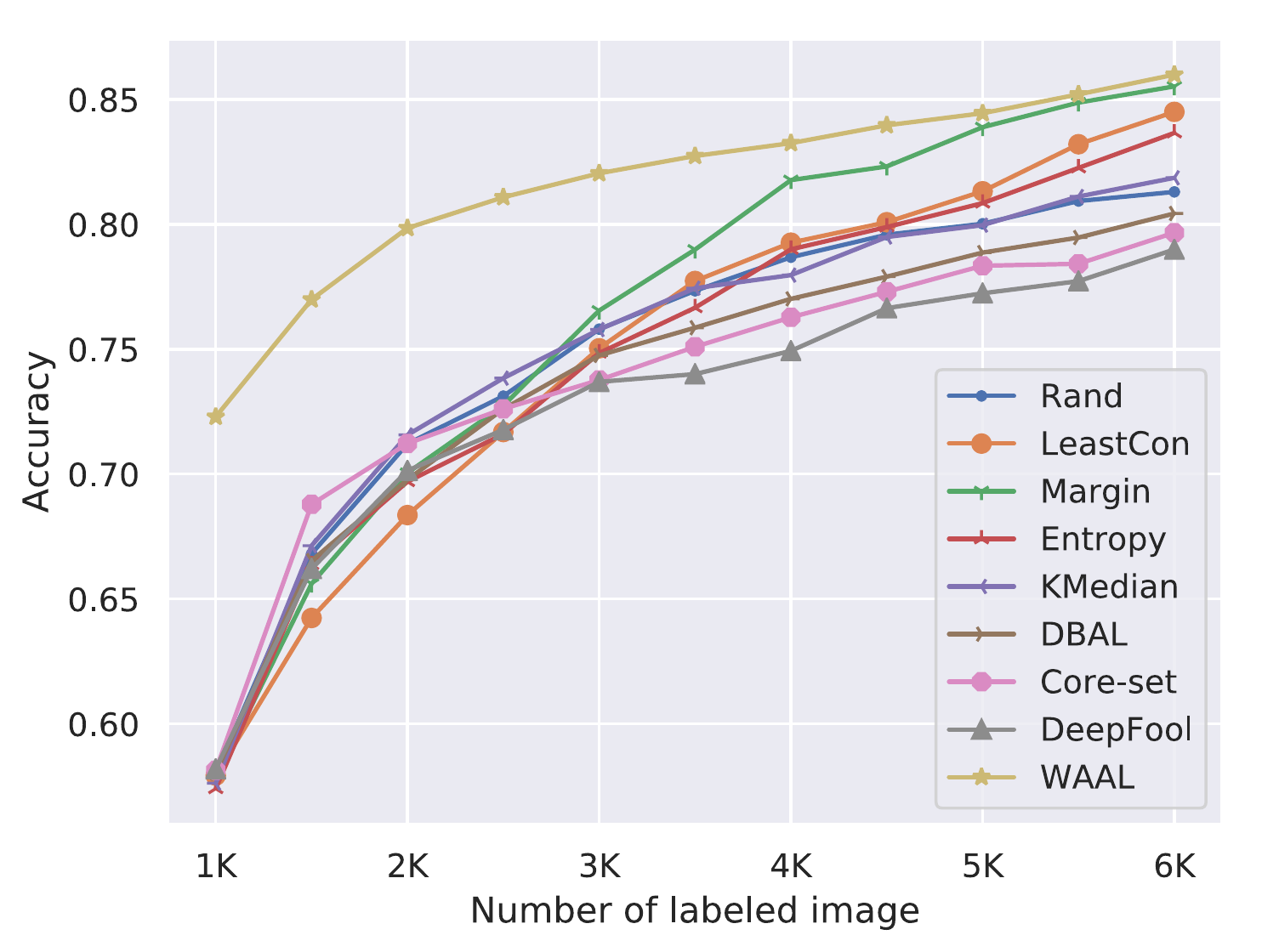}
   \caption{Fashion MNIST}
\end{subfigure}
\hfill
\begin{subfigure}{0.33\linewidth}
   \centering
   \includegraphics[width=\linewidth]{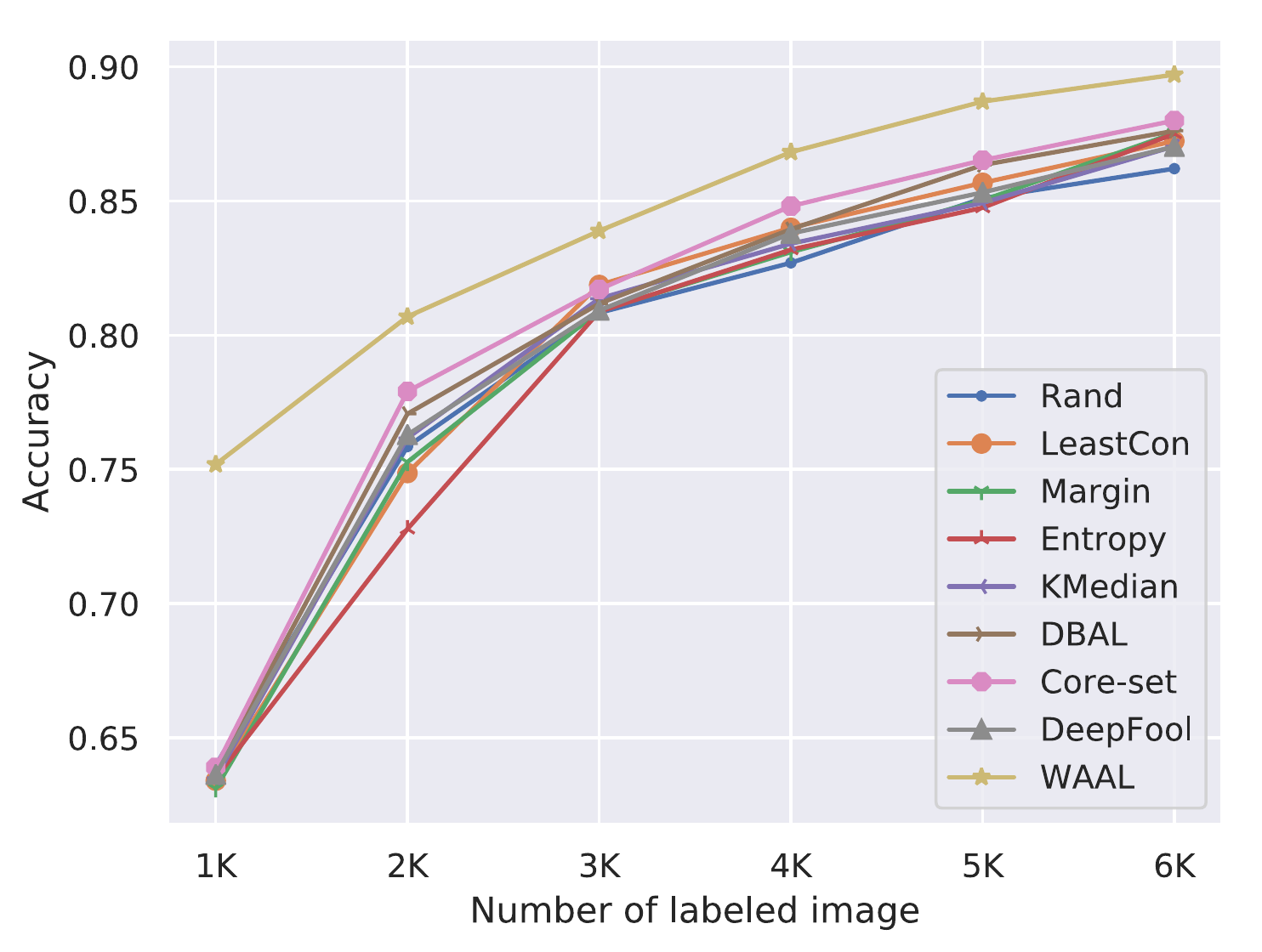}
   \caption{SVHN}
\end{subfigure}
\hfill
\centering
   \begin{subfigure}{0.33\linewidth}
   \centering
   \includegraphics[width=\linewidth]{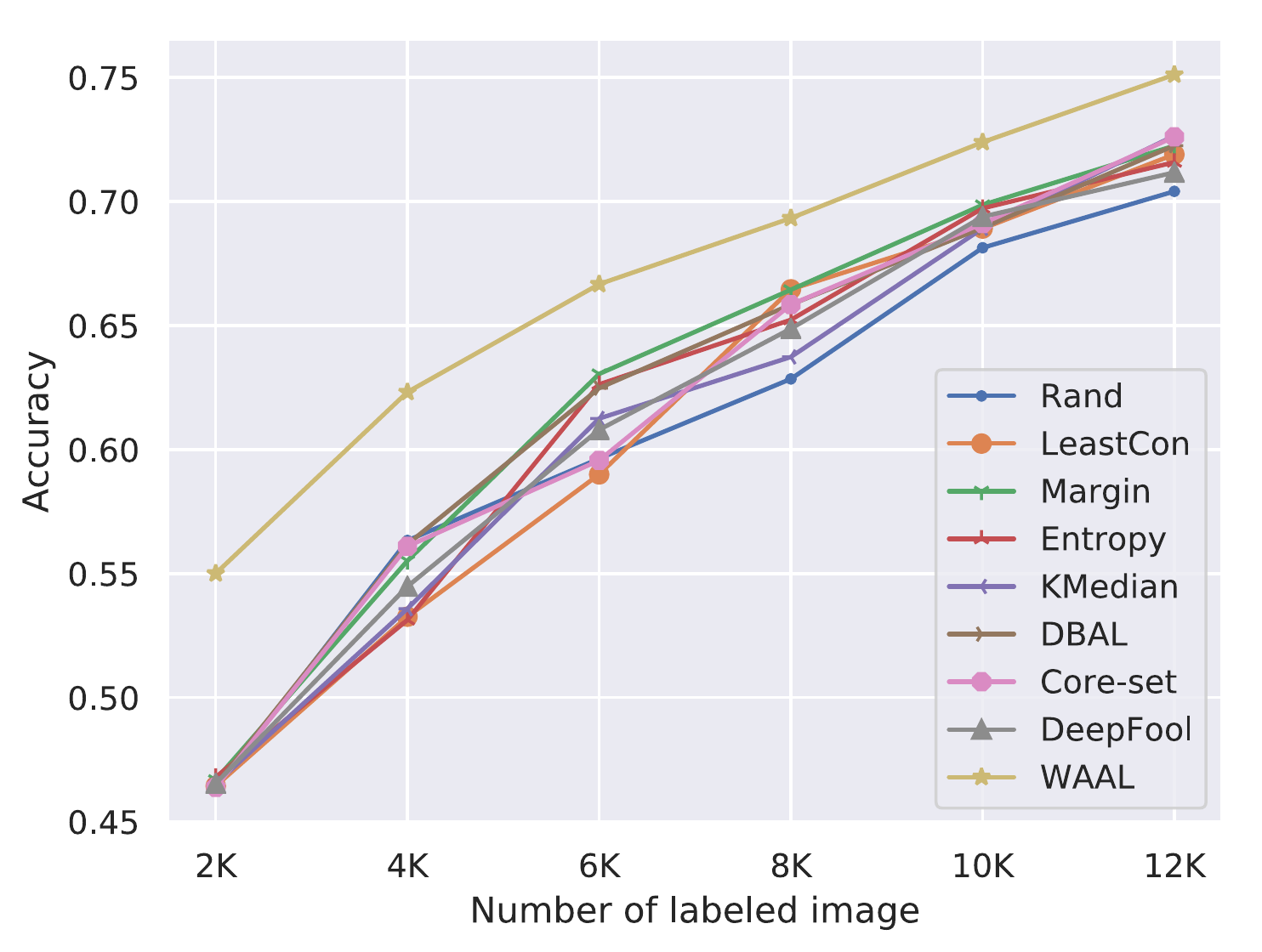}
   \caption{CIFAR-10}
\end{subfigure}
\caption{Empirical performance on Fashion MNIST, SVHN and CIFAR-10 over five repetitions.}
\label{fig:acc_result}
\end{figure*}

\begin{table*}[!t]
    \centering
    \begin{tabular}{@{}c|c|c|c|c|c|c|c|c@{}}
    \toprule
 Method & LeastCon  & Margin & Entropy  & $K$-Median & DBAL & Core-set & DeepFool & WAAL  \\ \midrule
 Time & $0.94$ & $0.95$ & $0.95$  & $33.98$ & $9.25$  & $45.88$   & $124.46$ & $1$ \\ \bottomrule
\end{tabular}
    \caption{Relative Average querying time, assuming the query time of WAAL as the unit.}
    \label{tab:query_time}
\end{table*}

\section{Experiments}
We start our experiments with a small initial labeled pool of the training set. The initial observation size and the budget size range from $1\%-5\%$ of the training dataset, depending on the task. Following Alg.~\ref{AALN_algo}, the selected batch will be annotated and added into the training set. Then the training process for the next iteration will be repeated on the new formed labeled and unlabeled set \emph{from scratch}.

We evaluate our proposed approach on three object recognition tasks, namely Fashion MNIST (image size: $28\times28$) \cite{xiao2017/online}, SVHN ($32\times32$) \cite{netzer2011reading}, CIFAR-10 ($32\times32$) \cite{krizhevsky2009learning}. For each task we split the whole data into training, validation and testing parts. We evaluate the performance of the proposed algorithm for image classification task by computing the prediction accuracy. We repeat all of the experiments 5 times and report the average value. The details of the experimental settings (dataset description, train/validation/test splitting, detailed implementations, hyper-parameter settings and choices) and additional experimental results are provided in the supplementary material.

\paragraph{Baselines} We compare the proposed approach with the following baselines: 1) Random sampling; 2) Least confidence \cite{culotta2005reducing}; 3) Smallest Margin \cite{scheffer2001active}; 4) Maximum-Entropy sampling \cite{settles2012active}; 5) $K$-Median approach \cite{sener2018active}: choosing the points to be labelled as the cluster centers of $K$-Median algorithm ; 6) Core-set approach \cite{sener2018active}; 7) Deep Bayesian AL (DBAL) \cite{gal2017deep}; and 8) DeepFoolAL \cite{mayer2018adversarial}.

\paragraph{Implementations} For the proposed approach, differing from baselines, we train the DNN from labeled and unlabeled data without data-augmentation. For the tasks on SVHN, CIFAR-10 we implement VGG16 \cite{simonyan2014very} and for task on Fashion MNIST we implement LeNet5 \cite{lecun1998gradient} as the feature extractor. On top of the feature extractor, we implement a  two-layer multi-layer perceptron (MLP) as the classifier and critic function. For all tasks, at each interaction we set the maximum training epoch as $80$. For each epoch, we feed the network with mini-batch of $64$ samples and adopt SGD with momentum \cite{sutskever2013importance} to optimize the network. We tune the hyper-parameter through grid search. In addition, in order to avoid over-training, we also adopt early stopping \cite{caruana2001overfitting} techniques during training.

\subsection{Results}
We demonstrate the empirical results in Fig.~\ref{fig:acc_result}. Exact numerical values and standard deviation are reported in the supplementary material. The proposed approach (WAAL) consistently outperforms all of the baselines during the interactions. We noticed that WAAL shows a large improvement ($>5\%$) in the initial training procedure since it efficiently constructs a good representation through leveraging the unlabeled data information. 
For the relatively simple input task Fashion MNIST, the simplest uncertainty query (Smallest Margin/Least Confidence) finally achieved almost the same level performance with WAAL under 6K labeled samples. 
Moreover, we observed that for the small or middle sized queried batch (0.5K-2K) in the relatively complex dataset (SVHN, CIFAR-10), the baselines show similar results in deep AL, which is coherent with previous observations \cite{DanielGissin2019,ash2019deep}. On the contrary, our proposed approach still shows a good improved empirical result, emphasizing the benefits of properly designing loss for considering the unlabeled data in the context of deep AL.

We also report the average query time for the baselines and proposed approach on SVHN dataset in Tab.~\ref{tab:query_time}. The results indicates that WAAL holds the same querying time level with the standard uncertainty based strategies since they are all \emph{end-to-end} strategies without knowing the internal information of the DNN. However some diversity based approaches such as Core-set and $K$-Median require the computation of the distance in the feature space and finally induce a much longer query time.

\subsection{Ablation Study: Advantage of Wasserstein Metric}
\begin{figure}[!t]
    \centering 
	\includegraphics[width=0.45\textwidth]{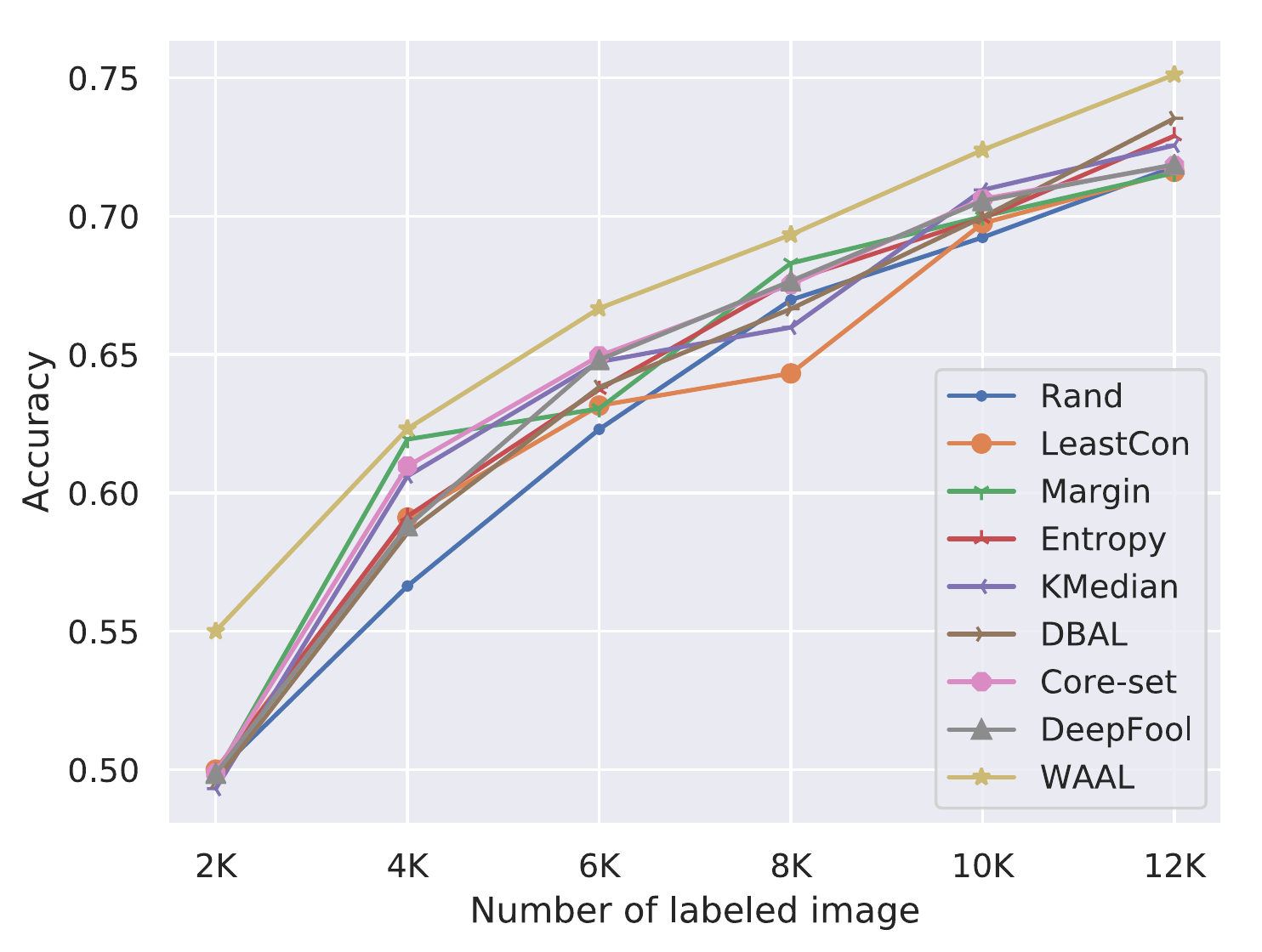}
	\caption{Ablation study in CIFAR-10: the baselines are all trained by leveraging the unlabeled information through $\calH$-divergence.}
\label{fig:abl}
\end{figure}

In this part, we empirically show the advantage of considering the Wasserstein distance by the ablation study. Specifically, for the whole baselines we adopt $\calH$-divergence based adversarial loss for training DNN. That is, we set a discriminator and we used the binary cross entropy (BCE) adversarial loss to discriminate the labeled and unlabeled data \cite{DanielGissin2019}. Then in the query we still apply the different baselines strategies to obtain the labels. We tested in the CIFAR-10 dataset and report the performances in Fig.~\ref{fig:abl}. Due to space limit, we present the brief introduction, exact numerical values, and more results in the supplementary material.

From the results, we observed that the gap between the initial training procedure has been reduced from about $8\%$ to $5\%$ because of introducing the adversarial based training. However, our proposed approach (WAAL) still consistently outperforms the baselines. The reason might be that the $\calH$-divergence based adversarial loss is not a good metric for the Deep AL as we formally analyzed before. The results indicate the practical potential of adopting the Wasserstein distance for the Deep AL problem.

\section{Related work}
\paragraph{Active learning} is a long term and promising research topic in theoretical and practical aspects. Here only the most related works are described and interested readers can find further details in the survey papers \cite{dasgupta2011two}, \cite{settles2012active} and \cite{hanneke2014theory}. In the context of deep AL, there are uncertainty and diversity based querying strategies. In the uncertainly based approach, we aim to find the difficult samples by using some heuristics, for example \cite{gal2017deep,beluch2018power,Haussmann2019active,pinsler2019bayesian} integrate uncertainty measure with a Bayesian deep neural network; \cite{DBLP:journals/corr/abs-1802-09841} adopt adversarial examples as the proxy distance in the margin based approach. In the diversity based query, \cite{geifman2017deep,sener2018active} apply the Core-set approach for selecting diverse data.

In addition, several works \cite{guo2008discriminative, wang2015querying} consider the mixture strategy of combing diversity and uncertainty, which are generally formulated as an iterative optimization problem in the AL, while they are not fit to the usual deep learning scenario.
\cite{bachman2017learning} combines the meta-learning with active learning for gradually choosing the proper acquisition function during the interactions. Several papers also adopt the idea of adversarial training (GAN) or variational approach (VAE), such as
\cite{zhu2017generative,mayer2018adversarial,tran2019bayesian}. However they generally simply plug in such generative techniques without analysing the superficiality in the AL.   

In the theoretical aspect, different approaches have been proposed such as exploiting the cluster structure through the hierarchical model, e.g.\ \cite{settles2012active}; or hypothesis space searching through estimating the disagreement region such as \cite{balcan2009agnostic}.  
These provable approaches generally provide a strong theoretical guarantee in AL, while they are highly computationally intractable in the DNN scenario. 

\paragraph{Distribution Matching} is an important research topic in the deep generative model \cite{goodfellow2014generative,kingma2013auto} and transfer learning \cite{ganin2016domain,shui2019principled}. In general, it aims at minimizing the statistical divergence between two distributions. 
In AL, some approaches implicitly connect with distribution matching.  \cite{DanielGissin2019} investigate $\calH$-divergence between the labeled and unlabeled dataset by constructing a binary classifier, while it has been shown that the $\calH$-divergence is not a proper metric for diversity in AL. \cite{chattopadhyay2013batch,wang2015querying,viering2017nuclear} adopted the Maximum Mean Discrepancy (MMD) metric by constructing a standard optimization problem, which also focuses on query strategy and is not scalable for the complex and large-scale dataset.

\section{Conclusion}
In this paper, we proposed a unified and principled method for both querying and training in the deep AL. We analyzed the theoretical insights from the intuition of modeling the interactive procedure in AL as distribution matching. Then we derived a new training loss for jointly learning hypothesis and query batch searching. We formulated the loss for DNN as a min-max optimization problem by leveraging the unlabeled data. As for the query for batch selection, it explicitly indicates the uncertainty-diversity trade-off. The results on different benchmarks showed a consistent better accuracy and faster efficient query strategy. The analytical and empirical results reaffirmed the benefits and potentials for reflecting on the unified principles for deep active learning. In the future, we want to 1) understand more general learning scenarios such as different distribution divergence metrics and its corresponding influences; 2) exploring other types of practical principles such as the auto-encoder based approach instead of adversarial training.

\section{Acknowledgements}
We would like to thank Mahdieh Abbasi and Qi Chen for the insightful discussions and feedback. 
We also appreciate Annette Schwerdtfeger for proofreading this manuscript.
C. Shui and C. Gagn\'e are funded by E Machine Learning, Mitacs, Prompt-Qu\'ebec, and NSERC-Canada. F. Zhou is supported by China Scholarship Council. B. Wang is supported by the Faculty of Science at the University of Western Ontario.

\newpage
\bibliographystyle{apalike}
\bibliography{ref}

\newpage
\onecolumn
\appendix

\begin{appendices}

\section{Theorem 1:\quad Proof}
\begin{theorem}
Supposing $\D$ is the data generation distribution and $\Q$ is the querying distribution, if the loss $\ell$ is symmetric, $L$-Lipschitz; $\forall h\in\calH$ is at most $H$-Lipschitz function and underlying labeling function $h^{\star}$ is $\phi(\lambda)$-$(\D,\Q)$ Joint Probabilistic Lipschitz, then the expected risk w.r.t. $\D$ can be upper bounded by:
\begin{equation*}
    R_\D(h) \leq R_{\Q}(h) + L(H+\lambda) W_1(\D,\Q)  +  L\phi(\lambda)
\end{equation*}
\end{theorem}

\subsection{Notations}
We define the hypothesis $h:\calX\to\calY=[0,1]$ and loss function $\ell: \calY\times\calY\to \R^{+}$, then the expected risk w.r.t. $\D$ is $R_{\D}(h) = \E_{x\sim\D} \ell(h(x),h^{\star}(x))$ and empirical risk $\hat{R}_{\D}(f) = \frac{1}{N}\sum_{i=1}^N \ell(h(x_i),y_i)$. We assume the loss $\ell$ is symmetric, $L$-Lipschitz and bounded by $M$.

\subsection{Transfer risk}

The first step is to bound the the gap $R_\D(h) - R_Q(h)$:
\begin{equation}
    \begin{split}
          R_\D(h) - R_Q(h) 
          & \leq |R_\D(h) - R_\Q(h)| = |E_{x\sim\D} \ell(h(x),h^{\star}(x)) - E_{x\sim \Q}\ell(h(x),h^{\star}(x))|\\
         & = |\int_{x\in\Omega}\ell(h(x),h^{\star}(x))d(\D-\Q)|
    \end{split}
    \label{risk_u1}
\end{equation}

From the Kantorovich - Rubinstein duality theorem  and combing Eq. (\ref{risk_u1}), for \textbf{any} distribution coupling $\gamma\in \Pi(\D,\Q)$, we have:
\begin{equation*}    
\begin{split}
    & = |\int_{\Omega\times\Omega} \big( \ell(h(x_\D),h^{\star}(x_\D))  -\ell(h(x_\Q),h^{\star}(x_\Q)) \big)  d \gamma(x_\D,x_\Q)|\\
    & \leq  \int_{\Omega\times\Omega} | \ell(h(x_\D),h^{\star}(x_\D)) -\ell(h(x_\Q),h^{\star}(x_\Q)) | d \gamma(x_\D,x_\Q) \\
    &  \leq \int_{\Omega\times\Omega}|\ell(h(x_\D),h^{\star}(x_\D))-\ell(h(x_\D),h^{\star}(x_\Q))| + |\ell(h(x_\D),h^{\star}(x_\Q))  -\ell(h(x_\Q),h^{\star}(x_\Q))|d \gamma(x_\D,x_\Q)
\end{split}
\end{equation*}

Since we assume $\ell$ is symmetric and $L$-Lipschitz, then we have:
\begin{equation}
\leq L \int_{\Omega\times\Omega}|h^{\star}(x_\D)-h^{\star}(x_Q)|d\gamma(x_\D,x_Q)  + L \int_{\Omega\times\Omega} |h(x_\D)-h(x_Q)|d\gamma(x_\D,x_Q)
    \label{risk_twoside}
\end{equation}

From Eq.(\ref{risk_twoside}) the risk gap is controlled by two terms, the property of labeling function and property of predictor. Moreover, we assume the learner is $H$-Lipschitz function, then we have:
\begin{equation*}
     \leq L \int_{\Omega\times\Omega}|h^{\star}(x_\D)-h^{\star}(x_\Q)|d\gamma(x_\D,x_\Q) 
        +  LH \int_{\Omega\times\Omega} ||x_\D-x_\Q||_2 d\gamma(x_\D,x_\Q) 
\end{equation*}

\paragraph{Labeling function assumption} As mentioned before, the goodness of the underlying labeling function decides the level of risk. \cite{urner2013plal} formalize the such a property as \emph{Probabilistic Lipschitz} condition in AL, in which relaxes the
condition of Lipschitzness condition and formalizes the intuition that \emph{under suitable
feature representation the probability of two close points having different labels is small}  \cite{urner2013probabilistic}. We adopt the joint probabilistic Lipschitz property, which is coherent with \cite{NIPS2017_6963}.
\begin{definition}
The labeling function $h^{\star}$ satisfies $\phi(\lambda)$-$(\D,\Q)$ Joint Probabilistic Lipschitz if $\mathrm{supp}(\Q)\subseteq\mathrm{supp}(\D)$ and for all $\lambda>0$:
\begin{equation}
     \Proba_{(x_\D,x_\Q)\sim\gamma}[|h^{\star}(x_\D)-h^{\star}(x_\Q)|>\lambda\|x_\D-x_\Q\|_2]\leq \phi(\lambda) 
     \label{eq_prob_lip}
\end{equation}
Where $\phi(\lambda)$ reflects the decay rate. \cite{urner2013plal} showed that the faster the decay of $\phi(\lambda)$ with $\lambda\to 0$, the nicer the distribution and the easier it is to learn the task.
\end{definition}

\noindent Combining with Eq.(\ref{eq_prob_lip}), the labeling function term can be decomposed and upper bounded by:
\begin{equation*}
    \begin{split}
    & \leq L \int_{\Omega\times\Omega} \mathbf{1} \{|h^{\star}(x_\D)-h^{\star}(x_\Q)| \leq \lambda \|x_\D-x_\Q\|_2\} |h^{\star}(x_\D)-h^{\star}(x_\Q)|d\gamma(x_\D,x_\Q) \\
& + L \int_{\Omega\times\Omega}  \mathbf{1} \{|h^{\star}(x_\D)-h^{\star}(x_\Q) > \lambda \|x_\D-x_\Q\|_2\} |h^{\star}(x_\D)-h^{\star}(x_Q)|d\gamma(x_\D,x_\Q) \\
& \leq L \lambda \int_{\Omega\times\Omega} \|x_\D-x_\Q\|_2 d\gamma(x_\D,x_\Q) + L\phi(\lambda)
    \end{split}
\end{equation*}
The first term is upper bounded through the probability of this event at most $1$ and second term adopts the definition of Joint Probabilistic Lipschitz with restricting the output space $h^{\star}(\cdot)\in[0,1]$. Plugging in the aforementioned results, we have:
\begin{equation*}
    \leq L(H+\lambda)  \int_{\Omega\times\Omega} \|x_\D-x_\Q\|_2 d\gamma(x_\D,x_\Q) + L\phi(\lambda)
\end{equation*}
Since this inequality satisfies with any distribution coupling $\gamma$, then it is also satisfies with the optimal coupling, w.r.t. the Wasserstein-1 distance with the cost function $\ell_2$ distance: $\|\cdot\|_2$. Then we have:
\begin{equation*}
    R_\D(h)-R_\Q(h) \leq L(H+\lambda) \inf_{\gamma} \int_{\Omega\times\Omega} \|x_\D-x_\Q\|_2 d\gamma(x_\D,x_\Q) +  L\phi(\lambda)
\end{equation*}
Finally we can derive:
\begin{equation}
    R_\D(h) \leq R_{\Q}(h) + L(H+\lambda) W_1(\D,\Q)  +  L\phi(\lambda)
    \label{con_1}
\end{equation}

\section{Corollary 1: Proof}
\subsection{Basic statistical learning theory}
According to the standard statistical learning theory such as \cite{mohri2018foundations}, w.h.p. $1-\delta/2$, $\forall h\in\calH$ we have:
\begin{equation}
    R_{\D}(h) \leq \hat{R}_{\D}(h) + 2L\Rad_N(h) + \kappa_1(\delta,N)
    \label{stl_error}
\end{equation}
Where $\Rad_N(h) = \E_{S\sim\D^{N}}\E_{\sigma_{1}^{N}}[\sup_{h} \frac{1}{N}\sum_{i=1}^N \sigma_i h(x_i)]$ is the expected Rademacher complexity with $\Rad_N(h) = \mathcal{O}(\sqrt{\frac{1}{N}})$, and $\kappa_1(\delta,N)= \mathcal{O}(\sqrt{\frac{M\log(2/\delta)}{N}})$ is the confidence term. \\
In the Active learning, the goal is to control the generalization error w.r.t. $(\D,h^{\star})$, thus from Eq.\ref{stl_error} we have: 
\begin{equation*}
    R_\D(h) \leq (R_\D(h) - R_{\Q}(h)) + \hat{R}_{\Q}(h) + 2L \Rad_{N_q}(h) + \kappa_1(\delta,N_q)
\end{equation*}
Combining with Eq.(\ref{con_1}), we have
\begin{equation*}
    R_\D(h) \leq \hat{R}_{\Q}(h) +  L(H+\lambda) W_1(\D,\Q)  +  L\phi(\lambda) + 2L \Rad_{N_q}(h) + \kappa_1(\delta,N_q)
\end{equation*}
In general we have finite observations (Supposing we have the sample i.i.d. sampled from query distribution $\Q$) with Dirac distribution: $\hat{D}=\frac{1}{N}\sum_{i=1}^N \delta\{x_{\D}^i\}$ and $\hat{Q}=\frac{1}{N_q}\sum_{i=1}^{N_q} \delta\{x_{\Q}^i\}$ with $N_{q}\leq N$. 
Several recent works show the concentration bound between empirical and expected Wasserstein distance such as \cite{bolley2007quantitative,weed2017sharp}. We just adopt the conclusion from \cite{weed2017sharp} and apply to bound the empirical measures in Wasserstein-1 distance.

\begin{lemma} \cite{weed2017sharp} [Definition 3,4] Given a measure $\mu$ on $X$, the $(\epsilon,\tau)$-covering number on a given set $S\subseteq X$ is:
$$\mathcal{N}_{\epsilon}(\mu,\tau) := \inf \{ \mathcal{N}_{\epsilon}(S): \mu(S)\geq 1-\tau \} $$
and the $(\epsilon,\mu)$-dimension is:
$$d_{\epsilon}(\mu,\tau) := \frac{\log \mathcal{N}_{\epsilon}(\mu,\epsilon)}{-\log\epsilon} $$
Then the upper Wasserstein-1 dimensions can be defined as:
$$ d_1^{\star}(\mu) = \inf\{s\in(2,+\infty): \lim\sup_{\epsilon \to 0} d_{\epsilon}(\mu,\epsilon^{-\frac{s}{s-2}})\leq s \}$$
\end{lemma}

\begin{lemma} \cite{weed2017sharp}[Theorem 1, Proposition 20] For $p = 1$ and $s \geq d_1^{\star}(\mu)$, there exists a positive constant $C$ with probability at least $1-\delta$, we have:
$$W_1(\mu,\hat{\mu}_N ) \leq  CN^{-1/s} + \sqrt{\frac{1}{2N}\log(\frac{1}{\delta})} $$
\end{lemma}
Since $s>2$ thus the convergence rate of Wasserstein distance is slower than $\mathcal{O}(N^{-1/2})$, also named as \emph{weak convergence}. 
Then according to the triangle inequality of Wasserstein-1 distance, we have:
\begin{equation}
    W_1(\D,\Q) \leq  W_1(\D,\hat{\D}) + W_1(\hat{\D},\Q)
   \leq W_1(\D,\hat{\D}) + W_1(\hat{\D},\hat{\Q}) + W_1(\hat{\Q},\Q) 
\end{equation}
Combing the conclusion with Lemma $2$, there exist some constants $(C_d,s_d)$ and $(C_q,s_q)$
we have with probability $1-\delta/2$:
\begin{equation}
         W_1(\D,\Q) \leq  W_1(\hat{\D},\hat{\Q}) + C_d N^{-1/s_d} + C_q N_q^{-1/s_q} 
        + \sqrt{\frac{1}{2}\log(\frac{2}{\delta})}(\sqrt{\frac{1}{N}} + \sqrt{\frac{1}{N_q}})
        \label{con_wa}
\end{equation}
Combining Eq.\ref{con_wa} and Eq.\ref{stl_error}, we have
\begin{equation*}
     R_\D(h) \leq \hat{R}_{\Q}(h) +  L(H+\lambda) W_1(\hat{\D},\hat{\Q})  +  L\phi(\lambda) + 2L \Rad_{N_q}(h) + \kappa(\delta,N,N_q)
\end{equation*}
Where $\kappa(\delta,N,N_q) = \mathcal{O}(N^{-1/s_d} + N_q^{-1/s_q} + \sqrt{\frac{\log(1/\delta)}{N}} + \sqrt{\frac{\log(1/\delta)}{N_q}})$

\section{Computing $\calH$-divergence and Wasserstein distance}
\subsection{$\calH$ divergence}
\paragraph{$d_{\calH}(\D_1,\D_3)$} We can discuss the discrepancy with different values of $p$, since $\D_3 \subseteq \D_1$ then we have $x_0 \in [a+b,2a-b]$:
\begin{enumerate}
    \item If $p\leq x_0-b $, then the area of mis-classification will be  $(2a-p)+2b $. If we select $p = x_0-b$, then the optimal mis-classification area will be $2a+2b-x_0 \geq 2a+2b-2a+b = 3b $   
    \item If $p\in [x_0-b, x_0+b]$, then the area of mis-classification will be $(2a-p)+(p-(x_0-b)) = 2a+b-x_0 \geq 2a+b-2a+b = 2b $
    \item If $p \geq x_0+b$, then the area of mis-classification will be $2b + \max(0,2a-p)$, if we select $p\geq 2a$, then the optimal mis-classification area will be $2b$.
\end{enumerate}
Then the minimal mis-classification area is $2b$, corresponding the optimal risk $\frac{b}{a+b}$, then $d_{\calH}(\D_1,\D_3) = \frac{b}{a+b}$

\paragraph{$d_{\calH}(\D_1,\D_2)$} We can discuss the discrepancy with different values of $p$, since $\D_2 \subseteq \D_1$, then we have $x_0 \in [a+b/2,2a-b/2]$:
\begin{enumerate}
    \item If $p\leq -x_0-b/2$, then the mis-classification area will be $a+\max(0,p+a)+2b$ with optimal value $2a+b/2-x_0+2b \geq a+2b$;
    \item If $p\in [-x_0-b/2, -x_0+b/2]$, then the mis-classification area will be $p-(-x_0-b/2) + (-a-p) + a + b = 
    x_0 + 3b/2 \geq a+2b$;
    \item If $p\in [-x_0+b/2, x_0-b/2]$, then the mis-classification area will be $b + a$;
    \item If $p\in [x_0-b/2, x_0+b/2]$, then the mis-classification area will be $b + p-(x_0-b/2) + (2a-p) = 2a+ 3b/2-x_0 \geq 2a+b/2-2a+b/2 + b = 2b $
    \item If $p \geq x_0 + b/2$, then the mis-classification area will be $2b + \max(0,2a-p) \geq 2b$
\end{enumerate}
Then the minimal mis-classification area is $2b$, corresponding the optimal risk $\frac{b}{a+b}$, then $d_{\calH}(\D_1,\D_2) = \frac{b}{a+b}$. \\
From the previous example $d_{\calH}(\D_1,\D_2) = d_{\calH}(\D_1,\D_3)  $, we show the $\calH$ divergence is not good metric for measuring the representative in the data space. Since we want the query distribution more diverse spread in the space, then $\calH$ may not be a good indicator.

\subsection{Wasserstein-1 distance} 
We can also estimate the distribution distance through Wasserstein-1 metric. From \cite{LarryWasserman19} we have:
\begin{equation*}
    W_1(P,Q) = \int_{0}^{1} |F^{-1}(z)-G^{-1}(z)|dz
\end{equation*}
where $F(z)$ and $G(z)$ is the CDF (cumulative density function) of distribution $P$ and $Q$, respectively.

\paragraph{CDF of $\D_1$, $\D_2$ and $\D_3$} 
\begin{enumerate}
    \item 
    \begin{equation*}
        F_1(z) = 
            \begin{cases}
                \frac{1}{2a}(z+2a) & -2a\leq z\leq -a\\
                \frac{1}{2}        &  -a \leq z \leq a \\
                \frac{1}{2a} z     &  a \leq z\leq 2a
            \end{cases}
    \end{equation*}
    
    \begin{equation*}
        F_1^{-1}(z) = 
            \begin{cases}
                2a(z-1)            & 0 \leq z < 1/2\\
                [-a,a]             & z = 1/2 \\
                2az                &  1/2 < z\leq 1
            \end{cases}
    \end{equation*}
    
    \item 
     \begin{equation*}
        F_2(z) = 
            \begin{cases}
                \frac{1}{2b}(z+x_0+b/2)       & -x_0-b/2 \leq z\leq -x_0+b/2\\
                \frac{1}{2}                   &  -x_0+b/2 \leq z \leq x_0-b/2 \\
                \frac{1}{2b} (z-x_0+3b/2)     &  x_0 - b/2 \leq z\leq x_0 + b/2
            \end{cases}
    \end{equation*}
    
    \begin{equation*}
        F_2^{-1}(z) = 
            \begin{cases}
                2bz-x_0-b/2                   & 0 \leq z < 1/2\\
                [-x_0+b/2, x_0-b/2]           & z= 1/2 \\
                2bz+x_0-3b/2                  & 1/2 < z\leq 1
            \end{cases}
    \end{equation*}
    
    \item 
    \begin{equation*}
        F_3(z) = \frac{1}{2b}(z-x_0+b) ~~ z\in[x_0-b,x_0+b]
    \end{equation*}
    
    \begin{equation*}
        F_3^{-1}(z) = 2bz+x_0-b ~~ z\in[0,1]
    \end{equation*}

\end{enumerate}

\paragraph{Computing $W_1(\D_1,\D_2)$} 
According to the definition, we can compute 
$$W_1(\D_1,\D_2) = \int_{0}^{1/2}|2a(z-1)-2bz-x_0-\frac{b}{2}|dz + \int_{1/2}^{1}|2az-2bz-x_0+\frac{3}{2}b|dz$$ 
We firstly compute $\int_{0}^{1/2}|2a(z-1)-2bz-x_0-\frac{b}{2}|dz $, since $2a(z-1)-2bz-x_0-\frac{b}{2}<0$ for $z\in [0,1/2]$ (since $-a-b-x_0-b/2<0$. Then we have:
\begin{equation*}
    \begin{split}
        \int_{0}^{1/2}|2a(z-1)-2bz-x_0-\frac{b}{2}|dz & = \int_{0}^{1/2} \{ -2a(z-1)+2bz+x_0+\frac{b}{2} \} dz \\
        & = \frac{3}{4}a + \frac{1}{2}x_0 + \frac{1}{2}b
    \end{split}
\end{equation*}
Then we compute the second part:
\begin{equation*}
    \begin{split}
       & \int_{1/2}^{1}|2az-2bz-x_0+\frac{3}{2}b|dz \\
       & = \int_{1/2}^{z_0} \{ (x_0-\frac{3}{2}b) - 2(a-b)z \} dz +  \int_{z_0}^{1} \{ 2(a-b)z-x_0+\frac{3}{2}b \} dz \\
       & = \frac{1}{2(a-b)}(x_0-\frac{3}{2}b)^2 - \frac{3}{2}(x_0-\frac{3}{2}b) + \frac{3}{4}(a-b)
    \end{split}
\end{equation*}
with $z_0 = \frac{x_0-3b/2}{2(a-b)}$. Therefore we can compute the wasserstein-1 distance between distribution $\D_1$ and $\D_2$:
\begin{equation*}
    = \frac{1}{2(a-b)}(x_0-\frac{3}{2}b)^2-x_0+2b+\frac{3}{2}a
\end{equation*}
With $x_0 \in [a+b/2,2a-b/2]$. If we take $x_0 = 2a-b/2$, we can get the maximum:
$$\max_{x_0}W_1(\D_1,\D_2) = \frac{3}{2}a - \frac{b}{2} $$

\paragraph{Computing $W_1(\D_1,\D_3)$}
According to definition, we can compute 
$$W_1(\D_1,\D_3) = \int_{0}^{1/2} |2a(z-1)-2bz-x_0+b|dz + \int_{1/2}^{1} |2az-2bz-x_0+b|dz $$
We firstly compute $ \int_{0}^{1/2} |2a(z-1)-2bz-x_0+b|dz$, since $2(a-b)z-2a-x_0+b \leq 0$ for $z\in [0,1/2]$.
(easy to verify: $2(a-b)z-2a-x_0+b \leq (a-b)-2a-x_0+b = -a-x_0 < 0$), then
\begin{equation*}
    \begin{split}
        & \int_{0}^{1/2} |2a(z-1)-2bz-x_0+b|dz =  \int_{0}^{1/2} (x_0+2a-b) -2(a-b)z dz \\
        & = \frac{1}{2}(x_0+2a-b) - \frac{1}{4}(a-b) = \frac{3}{4}a +x_0-\frac{1}{4}b
    \end{split}
\end{equation*}
Then we compute the second term $ \int_{1/2}^{1} |2az-2bz-x_0+b|dz$, we define
$z_0 = \frac{x_0-b}{2(a-b)}$ and we can verify that $z_0 \in [1/2,1]$, then this term can be decomposed as 
we can rewrite it as:
$$\int_{1/2}^{z_0} -2(a-b)z + (x_0-b) dz + \int_{z_0}^{1} 2(a-b)z -(x_0-b) dz $$
$$ = \frac{(x_0-b)^2}{2(a-b)} -\frac{3}{2}(x_0-b) + \frac{5}{4}(a-b) $$
Then $W_1(\D_1,\D_3) = \frac{(x_0-b)^2}{2(a-b)} -\frac{3}{2}(x_0-b) + \frac{5}{4}(a-b) + \frac{1}{2}x_0+ \frac{3}{4}a -\frac{b}{4}= \frac{1}{2(a-b)}(x_0-b)^2 -x_0 + 2a = \frac{1}{2(a-b)}(x_0-b)^2-x_0+2a$ since $x_0 \in [a+b, 2a-b]$, then we have:
$$\min_{x_0} W_1(\D_1,\D_3) = \frac{(x_0-b)^2}{2(a-b)} -(a+b) +2a = \frac{a^2}{2(a-b)}+a-b$$

We can verify: $\frac{a^2}{2(a-b)}+a-b > \frac{3}{2}a -\frac{b}{2}$ when $a>b$, then we have:
$$ \min_{x_0} W_1(\D_1,\D_3) > \max_{x_0}W_1(\D_1,\D_2)$$
which means in Wasserstein-1 distance metric, the diversity of two distribution can be much better measured.

\section{Developing loss in deep batch active learning}
We have the original loss:
\begin{equation}
    \min_{\btheta^f,\btheta^h,\hat{B}} \max_{\btheta^{d}}~ \E_{(x,y)\sim \hat{L}\cup\hat{B}} \ell(h(x,y)) + \mu  \big( \E_{x\sim \hat{\D}} [g(x)]- \E_{x\sim \hat{L}\cup\hat{B}} [g(x)] \big).
\end{equation}

Since $\hat{L}$, $\hat{B}$ and $\hat{\D}$ are Dirac distributions, then we have:
\begin{align*}
    & \frac{1}{L+B}\sum_{(x,y)\in\hat{L}\cup\hat{B}} \ell(h(x,y)) + \mu \big( \frac{1}{L+U}\sum_{x\in\hat{D}} g(x) - \frac{1}{L+B}\sum_{x\in\hat{L}\cup\hat{B}} g(x) \big)  \\
    & = \frac{1}{L+B}\sum_{(x,y)\in\hat{L}} \ell(h(x,y)) + \frac{1}{L+B}\sum_{(x,y^{?})\in\hat{B}} \ell(h(x,y^{?})) \\
    & + \mu\big( \frac{1}{L+U}\sum_{x\in\hat{L}} g(x) +   \frac{1}{L+U}\sum_{x\in\hat{U}} g(x) - \frac{1}{L+B}\sum_{x\in\hat{L}} g(x) -  \frac{1}{L+B}\sum_{x\in\hat{B}} g(x) \big)\\
    & = \frac{1}{L+B}\sum_{(x,y)\in\hat{L}} \ell(h(x,y)) + \frac{1}{L+B}\sum_{(x,y^{?})\in\hat{B}} \ell(h(x,y^{?})) \\
    & + \mu \big(\frac{1}{L+U}\sum_{x\in\hat{U}} g(x) - (\frac{1}{L+B}-\frac{1}{L+U})\sum_{x\in\hat{L}} g(x) \big) -  \frac{\mu}{L+B}\sum_{x\in\hat{B}} g(x)\\
    & = \Big(\underbrace{\frac{1}{L+B}\sum_{(x,y)\in\hat{L}} \ell(h(x,y))+ \mu \big(\frac{1}{L+U}\sum_{x\in\hat{U}} g(x) - (\frac{1}{L+B}-\frac{1}{L+U})\sum_{x\in\hat{L}} g(x) \big)  }_\text{Training Stage}\Big) \\
   & +  \Big(\underbrace{\frac{1}{L+B}\sum_{(x,y^{?})\in\hat{B}} \ell(h(x,y^{?})) -  \frac{\mu}{L+B}\sum_{x\in\hat{B}} g(x)}_\text{Querying Stage} \Big)
\end{align*}
We note that $x\in\hat{D}$ means enumerating all samples from the observations (empirical distribution). 

\section{Redundancy trick: Computation}
\begin{align*}
    & \mu \big(\frac{1}{L+U}\sum_{x\in\hat{U}}~ g(x) - (\frac{1}{L+B}-\frac{1}{L+U})\sum_{x\in\hat{L}} g(x) \big)\\
 & = \mu \big(\frac{\gamma}{1+\gamma} \frac{1}{U}\sum_{x\in\hat{U}}~ g(x)- (\frac{1}{1+\alpha} - \frac{1}{1+\gamma})\frac{1}{L}\sum_{x\in\hat{L}}g(x) \big)\\
 & = \mu^{\prime} \big(\frac{1}{U}\sum_{x\in\hat{U}}~g(x)- \frac{1}{\gamma}(\frac{1+\gamma}{1+\alpha} - 1)\frac{1}{L}\sum_{x\in\hat{L}}g(x) \big)\\ 
 & = \mu^{\prime} \big(\frac{1}{U}\sum_{x\in\hat{U}}~ g(x)- \frac{1}{\gamma}\frac{\gamma-\alpha}{1+\alpha} \frac{1}{L}\sum_{x\in\hat{L}}g(x) \big) 
\end{align*}

\section{Uniform Output Arrives the Minimal loss}
For the abuse of notation, we suppose the output of classifier $h(x,\cdot) = [p_1,\dots,p_K]\equiv \mathbf{p}$ with $p_i>0$ and $\sum_{i=1}^K p_i = 1$. Then we tried to minimize
\begin{equation*}
    \min_{\mathbf{p}} \sum_{i=1}^K -\log p_i
\end{equation*}
By applying the Lagrange Multiplier approach, we have 
\begin{align*}
    \min_{\mathbf{p},\lambda>0}  \sum_{i=1}^K -\log p_i + \lambda (\sum_{i=1}^K p_i -1)
\end{align*}
Then we do the partial derivative w.r.t. $p_i$, then we have $\forall i$:
$$ \frac{-1}{p_i} + \lambda = 0 ~\to~ p_i = \frac{1}{\lambda} $$
Given $\sum_{i=1}^K p_i= 1$, then we can compute $p_i = \frac{1}{K}$ arriving the minimial, i.e the uniform distribution.

\section{Experiments}
\subsection{Dataset Descriptions}
\begin{table}[!h]
\centering
\resizebox{0.8\textwidth}{!}{
\begin{tabular}{c|c|c|c|c|c|c}
Dataset & \#Classes & Train + Validation & Test & Initially labelled & Query size & Image size \\ \hline
Fashion-MNIST \cite{xiao2017/online} & 10 & 40K + 20K & 10K & 1K & 500      & $28\times28$ \\\hline
SVHN  \cite{netzer2011reading}        & 10 & 40K + 33K & 26K & 1K & 1K       & $32\times32$ \\\hline
CIFAR10  \cite{krizhevsky2009learning}     & 10 & 45K + 5K  & 10K & 2K & 2K       & $32\times32$ \\\hline
STL10* \cite{coates2011analysis} & 10 &  8K + 1K  & 4K &  0.5K & 0.5K   & $96\times96$
\end{tabular}}
\caption{Dataset descriptions}
\label{tab:data_details}
\end{table}
*We used a variant instead of the original STL10 dataset with arranging the training size to 8K (each class 800) and validation 1K and test 4K. We do not use the unlabeled dataset in our training or test procedure.

\subsection{Implementation details}
\paragraph{FashionMNIST} For the FashionMNIST dataset, we adopted the LeNet5 as feature extractor, then we used two-layer MLPs for the classification (320-50-relu-dropout-10) and critic function (320-50-relu-dropout-1-sigmoid). 
\paragraph{SVHN, CIFAR10} We adopt the VGG16 with batch normalization as feature extractor. then we used two-layer MLPs for the classification (512-50-relu-dropout-10) and critic function (512-50-relu-dropout-1-sigmoid).

\paragraph{STL10}  We adopt the VGG16 with batch normalization as feature extractor. then we used two-layer MLPs for the classification (4096-100-relu-dropout-10) and critic function (4096-100-relu-dropout-1-sigmoid).

\subsection{Hyper-parameter setting}
\begin{table}[!h]
\centering
\resizebox{0.8\textwidth}{!}{
\begin{tabular}{c|c|c|c|c|c|c|c}
Dataset       & lr & Momentum & Mini-Batch size & $\mu$  & Selection coefficient & Mixture coefficient**\\ \hline
Fashion-MNIST & 0.01* & 0.5 & 64 & 1e-2 & 5 & 0.5       \\\hline
SVHN          & 0.01* & 0.5 & 64 & 1e-2 & 5 & 0.5       \\\hline
CIFAR10       & 0.01* & 0.3 & 64 & 1e-2 & 10& 0.5        \\\hline
STL10         & 0.01* & 0.3 & 64 & 1e-3 & 10& 0.5  
\end{tabular}}
\caption{Hyper-parameter setting}
\label{tab:hyperpara}
\end{table}
* We set the initial learning rate as 0.01, then at 50\% epoch we decay to 1e-3, after 75\% epoch we decay to 1e-4.

** The mixture coefficient means the convex combination coefficient in the two uncertainty based approach.

\subsection{Detailed results with numerical values}
We report the accuracy in the form of percentage ($\%$), showing in Tab. \ref{tab:fmnist}, \ref{tab:svhn}, \ref{tab:cifar10}, \ref{tab:stl10}.

\begin{table*}[!t]
    \centering
    \resizebox{1.05\textwidth}{!}{
    \begin{tabular}{@{}c|c|c|c|c|c|c|c|c|c@{}}
    \toprule
      & Random   & LeastCon  & Margin & Entropy  & KMedian & DBAL & Core-set & DeepFool & WAAL  \\ \midrule
 1K & 58.03$\pm$2.81 & 57.93$\pm$1.62 & 57.81$\pm$2.19 & 57.40$\pm$1.75 & 57.62$\pm$2.5 & 58.01$\pm$2.75 & 58.14$\pm$2.19  & 58.19$\pm$2.4  &  72.29$\pm$1.16 \\\hline
 1.5K   & 66.81$\pm$1.02 & 64.24$\pm$2.49 & 65.61$\pm$ 2.5 & 66.37$\pm$0.62 & 67.13$\pm$2.87 & 66.53$\pm$2.5 & 68.79$\pm$1.99  & 66.21$\pm$1.78  &  76.99$\pm$1.05  \\\hline
 2K & 71.21$\pm$2.35 & 68.36$\pm$1.09 & 70.05$\pm$2.77 & 69.70$\pm$0.88 & 71.57$\pm$0.79 & 69.77$\pm$0.93 & 71.22$\pm$1.38  & 70.14$\pm$1.32  &  79.85$\pm$0.49  \\\hline
 2.5K   & 73.12$\pm$2.1 & 71.68$\pm$1.67 & 72.74$\pm$1.55 & 71.60$\pm$1.42 & 73.84$\pm$0.98 & 72.60$\pm$0.60 & 72.61$\pm$1.16  & 71.77$\pm$1.49  &  81.08$\pm$0.68  \\\hline
 3K & 75.80$\pm$0.64 & 75.03$\pm$1.56 & 76.55$\pm$1.01 & 74.84$\pm$1.29 & 75.79$\pm$0.44 & 74.75$\pm$1.04 & 73.77$\pm$1.74  & 73.69$\pm$1.21  &  82.04$\pm$0.58  \\\hline
 3.5K   & 77.34$\pm$0.67 & 77.73$\pm$1.04 & 78.99$\pm$1.11 & 76.66$\pm$1.26 & 77.44$\pm$0.97 & 75.86$\pm$1.02 & 75.10$\pm$1.11  & 74.00$\pm$0.71  &  82.74$\pm$0.79  \\\hline
 4K & 78.68$\pm$0.41 & 79.26$\pm$0.47 & 81.77$\pm$0.51 & 79.00$\pm$0.24 & 77.97$\pm$0.65 & 77.02$\pm$0.42 & 76.28 $\pm$0.98 & 74.93$\pm$2.05  &  83.25$\pm$0.62   \\\hline
 4.5K   & 79.58$\pm$0.47 & 80.08$\pm$0.82 & 82.32$\pm$0.47 & 79.89$\pm$0.78 & 79.49$\pm$0.7 & 77.90$\pm$0.58 & 77.30$\pm$0.61  & 76.64$\pm$0.97  &  83.96$\pm$0.54 \\\hline
 5K    & 80.02$\pm$0.45 & 81.32$\pm$0.64 & 83.89$\pm$0.84 & 80.85$\pm$0.87 & 79.97$\pm$0.59 & 78.87$\pm$0.58 & 78.34$\pm$0.37  & 77.24$\pm$0.69  &  84.45$\pm$0.45      \\\hline
 5.5K  & 80.93$\pm$0.33 & 83.21$\pm$0.42 & 84.87$\pm$0.18 & 82.26$\pm$0.77 & 81.11$\pm$0.41 & 79.47$\pm$2.9 & 78.42$\pm$0.66  & 77.72$\pm$0.57 & 85.20$\pm$0.44 \\\hline
 6K   & 81.30$\pm$0.25 &  84.50$\pm$0.73 & 85.52$\pm$0.27 & 83.66$\pm$0.98 & 81.86$\pm$0.6 & 80.43$\pm$0.76 & 79.66$\pm$0.34 & 78.99$\pm$0.33 &  85.99$\pm$0.43   \\\bottomrule
\end{tabular}}
    \caption{Result of FashionMNIST (Average $\pm$ std)}
    \label{tab:fmnist}
\end{table*}

\begin{table*}[!t]
    \centering
    \resizebox{1.05\textwidth}{!}{
    \begin{tabular}{@{}c|c|c|c|c|c|c|c|c|c@{}}
    \toprule
      & Random   & LeastCon  & Margin & Entropy  & KMedian & DBAL & Core-set & DeepFool & WAAL  \\ \midrule
 1K & 63.97$\pm$2.04 & 63.40$\pm$2.16 & 63.10$\pm$2.3 & 63.49$\pm$2.79 & 63.50$\pm$2.53 & 63.76$\pm$0.73 & 63.90$\pm$1.07  & 63.62$\pm$2.34  &  75.18$\pm$1.41 \\\hline
 2K   & 75.85$\pm$1.16 & 74.86$\pm$2.44 & 75.27$\pm$ 1.7 & 72.78$\pm$3.15 & 76.17$\pm$3.2 & 77.07$\pm$1.57 & 77.9$\pm$1.25  & 76.29$\pm$1.62  &  80.69$\pm$2.00  \\\hline
 3K & 80.83$\pm$ 1.04 & 81.87$\pm$ 	0.64 & 80.9$\pm$ 2.22 & 80.88$\pm$ 1.26 & 81.36$\pm$	1.29 & 81.17$\pm$ 1.72 & 81.7 $\pm$ 0.84  & 80.92$\pm$ 	0.79  &  83.89$\pm$ 2.08  \\\hline
 4K & 82.70$\pm$1.18 & 84.00$\pm$0.88 & 83.10$\pm$1.38 & 83.19$\pm$0.95 & 83.41$\pm$1.58 & 83.95$\pm$1.87 & 84.81$\pm$1.3  & 83.79$\pm$0.64  &  86.82$\pm$1.11  \\\hline
 5K & 85.10$\pm$0.73 & 85.68$\pm$0.94 & 85.02$\pm$1.1 & 84.75$\pm$0.83 & 84.93$\pm$0.94 & 86.34$\pm$1.1 & 86.52$\pm$0.95  & 85.32$\pm$0.58  &  88.71$\pm$1.08  \\\hline
 6K & 86.20$\pm$0.48 & 87.23$\pm$0.97 & 87.53$\pm$0.63 & 87.51$\pm$0.50 & 87.04$\pm$0.45 & 87.61$\pm$0.72 & 88.00 $\pm$0.44 & 87.02$\pm$0.64  &  89.71$\pm$0.83   \\\bottomrule
\end{tabular}}
    \caption{Result of SVHN (Average $\pm$ std)}
    \label{tab:svhn}
\end{table*}

\begin{table*}[!t]
    \centering
    \resizebox{1.05\textwidth}{!}{
    \begin{tabular}{@{}c|c|c|c|c|c|c|c|c|c@{}}
    \toprule
      & Random   & LeastCon  & Margin & Entropy  & KMedian & DBAL & Core-set & DeepFool & WAAL  \\ \midrule
 2K & 46.33$\pm$3.18 & 46.43$\pm$3.17 & 46.69$\pm$3.87 & 46.79$\pm$3.62 & 46.53$\pm$3.39 & 46.48$\pm$3.11 & 46.38$\pm$4.03  & 46.54$\pm$3.77  &  55.00$\pm$0.40 \\\hline
 4K   & 56.33$\pm$3.40 & 53.26$\pm$3.84 & 55.52$\pm$ 2.69 & 53.13$\pm$2.99 & 53.58$\pm$2.57 & 56.18$\pm$2.37 & 56.09$\pm$3.89  & 54.48$\pm$1.62  &  62.32$\pm$0.36  \\\hline
 6K & 59.63$\pm$ 4.17 & 59.00$\pm$ 2.19 & 63.05$\pm$ 1.78 & 62.63$\pm$ 1.29 & 61.25$\pm$1.76 & 62.48$\pm$ 1.38 & 59.56 $\pm$ 1.17  & 60.80$\pm$ 0.70  &  66.67$\pm$ 0.60  \\\hline
 8K & 62.85$\pm$3.37 & 66.46$\pm$1.33 & 66.44$\pm$1.85 & 65.23$\pm$1.89 & 63.73$\pm$1.34 & 65.84$\pm$0.78 & 65.84$\pm$1.27  & 64.87$\pm$1.98  &  69.33$\pm$1.47  \\\hline
 10K & 68.13$\pm$2.53 & 68.91$\pm$1.10 & 69.86$\pm$0.24 & 69.72$\pm$1.53 & 68.92$\pm$2.33 & 68.94$\pm$1.96 & 69.11$\pm$0.80  & 69.39$\pm$0.47  &  72.39$\pm$1.21  \\\hline
 12K & 70.41$\pm$1.02 & 71.90$\pm$1.35 & 72.25$\pm$0.68 & 71.58$\pm$0.77 & 72.65$\pm$0.64 & 72.25$\pm$1.24 & 72.60 $\pm$0.79 & 71.17$\pm$1.03  &  75.11$\pm$0.49   \\\bottomrule
\end{tabular}}
    \caption{Result of CIFAR10 (Average $\pm$ std)}
    \label{tab:cifar10}
\end{table*}

\begin{table*}[!t]
    \centering
    \resizebox{1.05\textwidth}{!}{
    \begin{tabular}{@{}c|c|c|c|c|c|c|c|c|c@{}}
    \toprule
      & Random   & LeastCon  & Margin & Entropy  & KMedian & DBAL & Core-set & DeepFool & WAAL  \\ \midrule
 0.5K & 41.78$\pm$2.42 & 41.69$\pm$3.22 & 41.81$\pm$2.27 & 41.12$\pm$1.67 & 41.24$\pm$1.41 & 41.30$\pm$1.45 & 41.41$\pm$2.30  & 41.82$\pm$2.67  &  47.01$\pm$1.09 \\\hline
 1K   & 48.24$\pm$1.37 & 47.05$\pm$1.42 & 46.7$\pm$0.85 & 46.38$\pm$2.31 & 46.45$\pm$1.11 & 47.45$\pm$3.71 & 47.58$\pm$2.06 & 45.15$\pm$0.74  & 52.47$\pm$1.62   \\\hline
 1.5K & 51.78$\pm$ 2.5 & 50.87$\pm$ 1.24 & 50.44$\pm$ 2.57 & 50.24$\pm$ 1.21 & 49.91$\pm$1.74 & 52.53$\pm$ 1.29 & 51.2 $\pm$ 1.63  & 48.64$\pm$ 	2.43  &  57.25$\pm$ 1.78  \\\hline
 2K & 56.52$\pm$1.78 & 56.25$\pm$1.58 & 55.54$\pm$1.09 & 55.15$\pm$2.13 & 54.92$\pm$2.19 & 57.54$\pm$1.70 & 58.13$\pm$1.57  & 54.26$\pm$2.40  &  60.08$\pm$1.63  \\\hline
 2.5K & 58.42$\pm$1.42 & 58.49$\pm$2.05 & 57.62$\pm$1.42 & 57.81$\pm$2.87 & 57.87$\pm$1.51 & 59.25$\pm$2.89 & 57.66$\pm$1.79  & 57.05$\pm$2.53  &  62.58$\pm$1.44  \\\hline
 3K & 61.13$\pm$1.67 & 60.80$\pm$2.64 & 59.42$\pm$1.49 & 60.88$\pm$0.72 & 60.00$\pm$0.65 & 62.11$\pm$1.65 & 61.02 $\pm$0.48 & 59.74$\pm$1.74  &  65.42$\pm$1.33   \\\bottomrule
\end{tabular}}
    \caption{Result of STL10 (Average $\pm$ std)}
    \label{tab:stl10}
\end{table*}

\section{Ablation study}
\begin{figure*}[!t]
\centering
   \begin{subfigure}{0.45\linewidth}
   \centering
   \includegraphics[width=\linewidth]{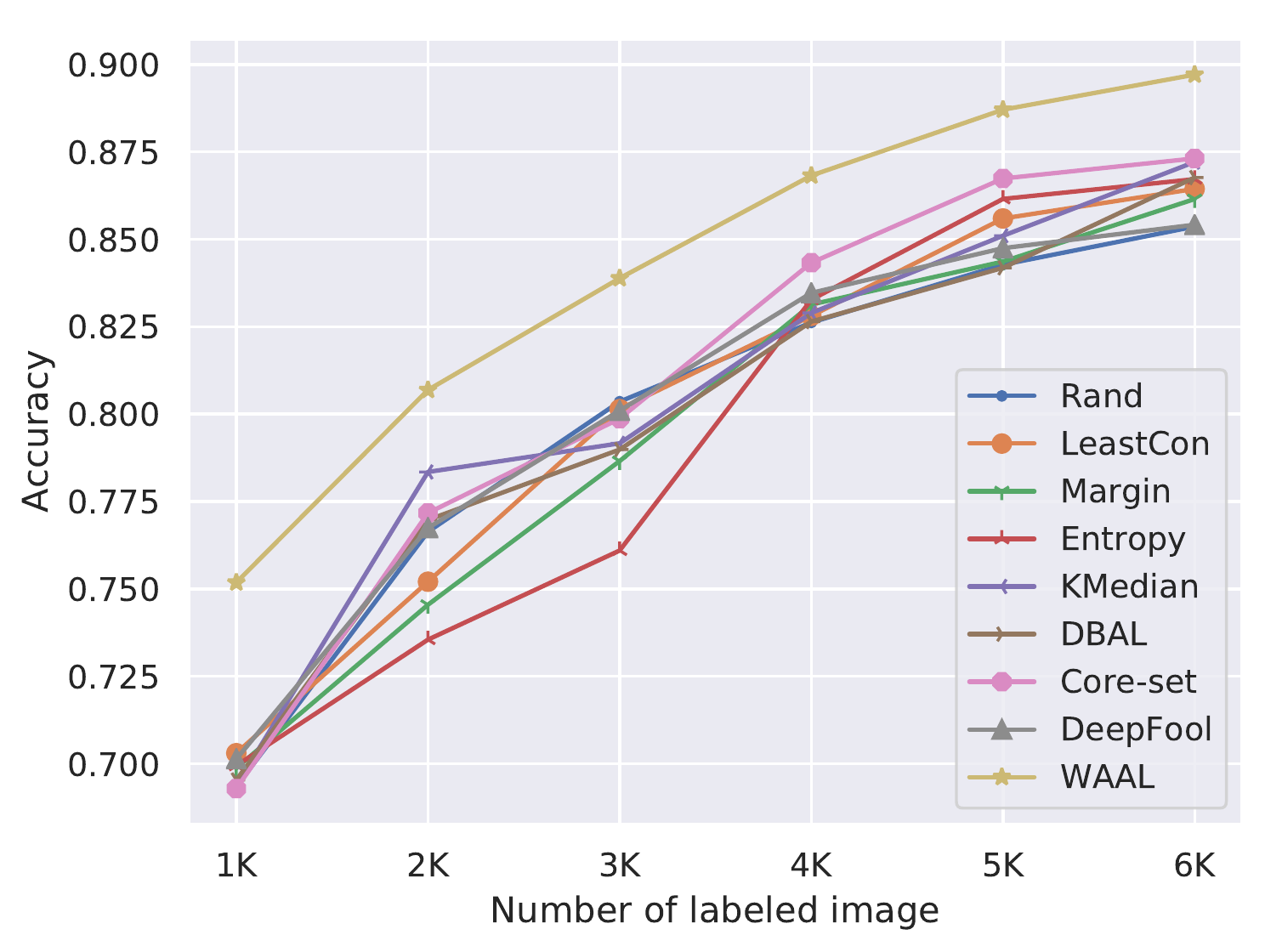}
   \caption{}
\end{subfigure}
\hfill
\begin{subfigure}{0.45\linewidth}
   \centering
   \includegraphics[width=\linewidth]{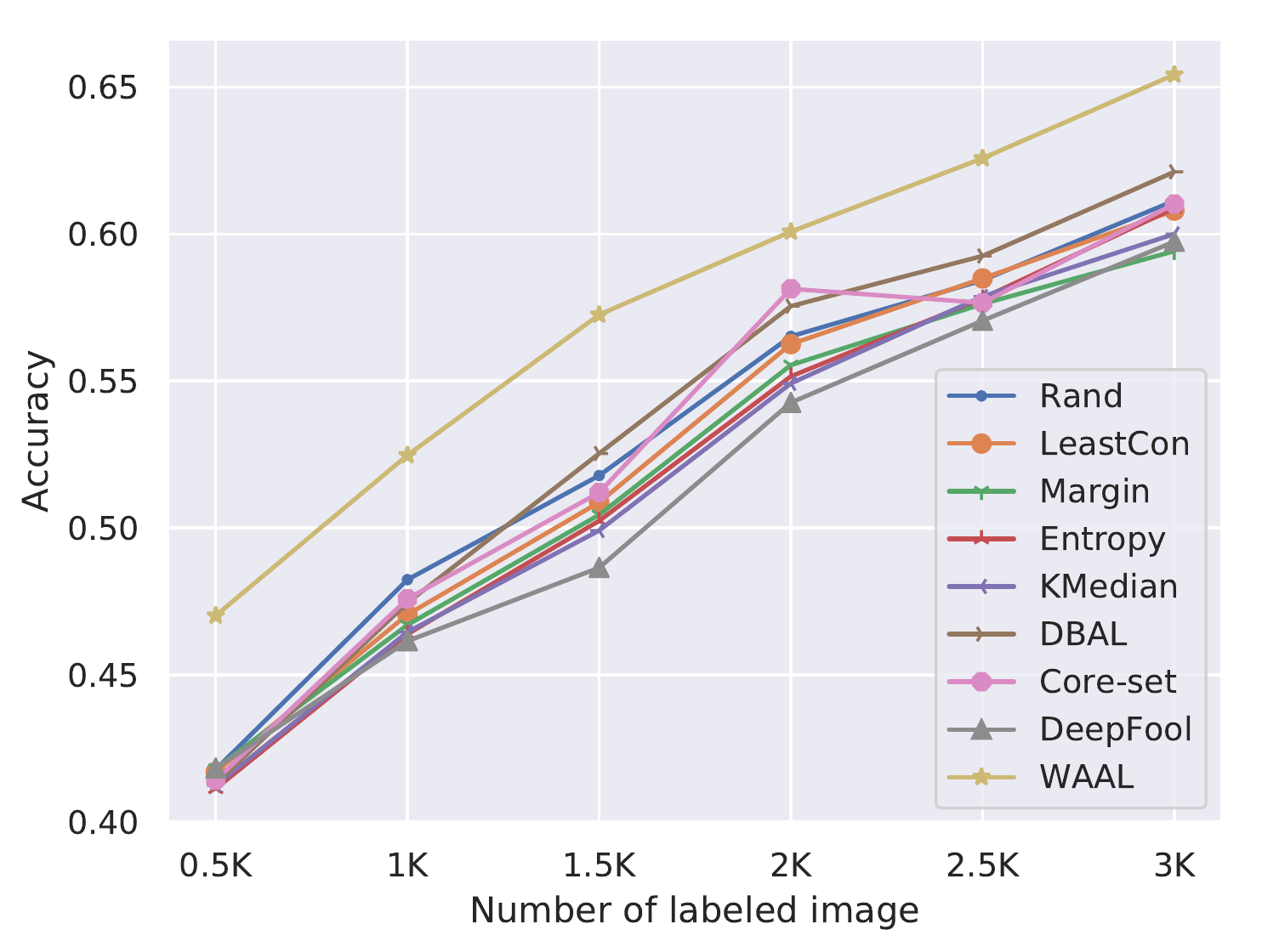}
   \caption{}
\end{subfigure}
\caption{(a) Ablation study in SVHN: the baselines are all trained by leveraging the unlabeled information through $\calH$-divergence; (b) Empirical performance on STL10, over five repetitions.}
\label{fig:acc_result_2}
\end{figure*}

In this part, we will conduct $\calH$-divergence based adversarial training for the parameters of DNN. 
\begin{equation}
    \min_{\btheta^f,\btheta^h}\max_{\btheta^{d}}~  \sum_{(x,y)\in\hat{L}} \ell(h(x,y)) - \mu \big(\sum_{x\in\hat{U}}\log(g(x)) + \sum_{x\in\hat{L}}\log(1-g(x))\big) 
    \label{H_train_loss}
\end{equation}
Where the $g$ is defined as the discriminator function \footnote{This notation is slightly different from the critic function \cite{arjovsky2017wasserstein}}. In the adversarial training, the discriminator parameter aims at discriminating the  empirical unlabeled and labeled data via the binary classification, while the feature extractor parameter aims at not being correctly classified. By this manner, the unlabeled dataset will be used for constructing a better feature representation in the adversarial training. As for the query part, we directly used baseline strategies. The numerical values will show in Tab. \ref{tab:abl_cifar10}. Moreover, we also evaluated the ablation study for the SVHN dataset, showing in Tab. \ref{tab:abl_svhn} and Fig. \ref{fig:acc_result_2}.

\begin{table*}[!t]
    \centering
    \resizebox{1.05\textwidth}{!}{
    \begin{tabular}{@{}c|c|c|c|c|c|c|c|c|c@{}}
    \toprule
      & Random   & LeastCon  & Margin & Entropy  & KMedian & DBAL & Core-set & DeepFool & WAAL  \\ \midrule
 2K & 49.85$\pm$0.32 & 50.00$\pm$1.81 & 49.8$\pm$2.28 & 49.92$\pm$1.08 & 49.31$\pm$2.76 & 49.58$\pm$1.05 & 49.87$\pm$4.03  & 49.85$\pm$1.36  &  55.00$\pm$0.40 \\\hline
 4K   & 56.63$\pm$3.27 & 59.11$\pm$0.85 & 61.93$\pm$ 2.12 & 59.15$\pm$0.41 & 60.6$\pm$0.72 & 58.55$\pm$1.99 & 60.97$\pm$1.62  & 58.80$\pm$2.59  &  62.32$\pm$0.36  \\\hline
 6K & 62.30$\pm$ 2.54 & 63.15$\pm$ 2.21 & 63.04$\pm$ 1.98 & 63.74$\pm$ 0.94 & 64.73$\pm$1.37 & 63.82$\pm$ 2.33 & 64.95 $\pm$ 1.66  & 64.80$\pm$ 1.4  &  66.67$\pm$ 0.60  \\\hline
 8K & 66.97$\pm$0.76 & 64.32$\pm$2.58 & 68.30$\pm$1.02 & 67.67$\pm$1.04 & 65.98$\pm$1.45 & 66.65$\pm$1.00 & 67.54$\pm$2.16  & 67.65$\pm$1.27  &  69.33$\pm$1.47  \\\hline
 10K & 69.23$\pm$1.97 & 69.74$\pm$2.52 & 69.98$\pm$0.25 & 69.92$\pm$1.17 & 70.95$\pm$1.93 & 69.96$\pm$1.74 & 70.62$\pm$0.74 & 70.55$\pm$0.80  &  72.39$\pm$1.21  \\\hline
 12K & 71.78$\pm$1.34 & 71.60$\pm$1.25 & 71.56$\pm$1.53 & 72.90$\pm$1.37 & 72.56$\pm$1.39 & 73.53$\pm$1.71 & 71.83 $\pm$1.20 & 71.86$\pm$0.33  &  75.11$\pm$0.49   \\\bottomrule
\end{tabular}}
    \caption{Ablation study of CIFAR10 (Average $\pm$ std)}
    \label{tab:abl_cifar10}
\end{table*}

\begin{table*}[!t]
    \centering
    \resizebox{1.05\textwidth}{!}{
    \begin{tabular}{@{}c|c|c|c|c|c|c|c|c|c@{}}
    \toprule
      & Random   & LeastCon  & Margin & Entropy  & KMedian & DBAL & Core-set & DeepFool & WAAL  \\ \midrule
 1K & 68.34$\pm$0.96 & 70.3$\pm$1.75 & 68.88$\pm$1.19 & 68.94$\pm$1.17 & 68.38$\pm$0.92 & 68.54$\pm$3.65 & 69.29$\pm$0.71  & 70.14$\pm$1.84  &  75.18$\pm$1.41 \\\hline
 2K   & 76.63$\pm$3.14 & 75.21$\pm$2.45 & 74.55$\pm$ 3.16 & 73.55$\pm$2.49 & 78.35$\pm$1.63 & 76.97$\pm$1.19 & 77.17$\pm$1.8 & 76.74$\pm$2.15  &  80.69$\pm$2.00  \\\hline
 3K & 80.36$\pm$ 0.46 & 80.14$\pm$1.66 & 78.66$\pm$ 1.54 & 76.10$\pm$ 1.46 & 79.16$\pm$	1.47 & 78.99$\pm$ 1.57 & 79.87 $\pm$ 0.33  & 80.10$\pm$ 	1.27  &  83.89$\pm$ 2.08  \\\hline
 4K & 82.62$\pm$1.15 & 82.81$\pm$0.66 & 83.13$\pm$1.01 & 83.27$\pm$0.18 & 82.89$\pm$0.73 & 82.65$\pm$1.61 & 84.33$\pm$0.72  & 83.47$\pm$0.74  &  86.82$\pm$1.11  \\\hline
 5K & 84.27$\pm$0.77 & 85.59$\pm$0.74 & 84.36$\pm$0.75 & 86.15$\pm$0.23 & 85.10$\pm$0.57 & 84.18$\pm$0.25 & 86.74$\pm$0.34  & 84.75$\pm$0.57  &  88.71$\pm$1.08  \\\hline
 6K & 85.36$\pm$0.36 & 86.44$\pm$0.93 & 86.15$\pm$0.89 & 86.72 $\pm$ 0.66 & 87.21$\pm$0.52 & 86.77$\pm$1.26 & 87.31$\pm$0.71 & 85.42 $\pm$0.55 &  89.71$\pm$0.83   \\\bottomrule
\end{tabular}}
    \caption{Ablation study of SVHN (Average $\pm$ std)}
    \label{tab:abl_svhn}
\end{table*}

\end{appendices}

\end{document}